%% file: main.tex
\documentclass[journal, a4paper]{ieeetj}
\usepackage{fontspec}
\usepackage{unicode-math}
\defaultfontfeatures{Ligatures=TeX}
\usepackage[english]{babel}
\usepackage{graphicx,color}
\usepackage{textcomp}
\usepackage[dvipsnames,table,rgb]{xcolor}
\usepackage[
    hidelinks,
    pdftitle = {MOSAIC: Modular Supervised Autonomy for Intelligent Coordination of Heterogeneous Robotic Teams},
    pdfauthor={},      %
    pdfsubject={},
    pdfkeywords={Autonomy, Multi-robot systems, Field robots, Space robotics}
    ]{hyperref}
\usepackage{csquotes}
\usepackage[capitalise]{cleveref}
\usepackage{array}
\usepackage{tabularx}
\newcolumntype{K}[1]{>{\RaggedRight\arraybackslash}p{#1}}
\usepackage{booktabs}   %
\usepackage{multirow}   %
\usepackage{siunitx}    %
\usepackage{lettrine}
\usepackage{ragged2e}    %
\usepackage{adjustbox}
\usepackage{xspace}
\makeatletter
\renewcommand*{\p@subsection}{\thesection.}
\renewcommand*{\p@subsubsection}{\thesection.\thesubsection.}
\renewcommand*{\p@paragraph}{\thesection.\thesubsection.\thesubsubsection.}
\renewcommand\paragraph{%
  \@startsection{paragraph}
    {4}{\parindent}
    {\z@}
    {-1em}%
    {\normalfont\normalsize\bfseries}
}
\makeatother

\usepackage[caption=false, font=footnotesize, labelfont=sf, textfont=sf]{subfig}

\hypersetup{pdfpagemode=UseNone}
\definecolor{MidGreen}{RGB}{60,176,67}
\definecolor{TfrBlue}{RGB}{13, 99, 212}

\newcommand\copyrighttext{%
	\footnotesize \textcopyright \the\year{} IEEE. Personal use of this material is permitted.
	Permission from IEEE must be obtained for all other uses, in any current or future
	media, including reprinting/republishing this material for advertising or promotional
	purposes, creating new collective works, for resale or redistribution to servers or
	lists, or reuse of any copyrighted component of this work in other works.}
\newcommand\copyrightnotice{%
	\begin{tikzpicture}[remember picture,overlay]
		\node[anchor=south,yshift=40pt,xshift=50pt] at (current page.south) {\fbox{\parbox{\dimexpr\textwidth-\fboxsep-\fboxrule\relax}{\copyrighttext}}};
	\end{tikzpicture}%
}

\hypersetup{
    colorlinks=true,
    linkcolor=TfrBlue,
    linkbordercolor=white,
    filecolor=magenta,
    citecolor=MidGreen,
    urlcolor=TfrBlue
}

\usepackage{tikz}
\usetikzlibrary{
    decorations.pathreplacing,
    fit,
    calc,
    positioning,
    shapes,
    arrows,
    backgrounds,
    arrows.meta,
    matrix,
    patterns,
    shapes.arrows,
    shapes.symbols
    }

\usepackage{pgfplots}
\pgfplotsset{compat=1.18} 
\usepackage{listings}
\usepackage{algorithm}
\usepackage{algpseudocode}
\usepackage{float}
\newfloat{algorithm}{t}{lop}
\usepackage{censor}
\usepackage{flushend}
\algdef{SE}%
[STRUCT]%
{Struct}%
{EndStruct}%
[1]%
{\textbf{struct} \textsc{#1}}%
{\textbf{end struct}}%

\def\BibTeX{{\rm B\kern-.05em{\sc i\kern-.025em b}\kern-.08em
    T\kern-.1667em\lower.7ex\hbox{E}\kern-.125emX}}
\AtBeginDocument{\definecolor{tmlcncolor}{cmyk}{0.93,0.59,0.15,0.02}\definecolor{NavyBlue}{RGB}{0,86,125}}
\usepackage[acronym,numberedsection=false,nomain]{glossaries}
\glsdisablehyper{}
\definecolor{c_rock_markers}{HTML}{FFA500}
\definecolor{c_sand_markers}{HTML}{0cb6e6}
\usepackage{orcidlink}

\input{glossary}

\usepackage[
    backend     = biber,
    style       = ieee,
    seconds     = true,
    natbib      = true,           %
    sorting     = none,           %
    maxbibnames = 6,           %
    bibencoding = utf8,
    maxcitenames=1,
    mincitenames=1
    ]{biblatex}
\DeclareNameAlias{author}{family-given}
\AtEveryBibitem{%
    \clearfield{keywords}%
    \clearfield{note}%
}
\AtEveryBibitem{%
  \ifboolexpr{
    test {\ifentrytype{inproceedings}}
    or
    test {\ifentrytype{proceedings}}
    or
    test {\ifentrytype{article}}
  }{%
    \clearfield{url}%
    \clearfield{urldate}%
    \clearfield{urlyear}%
    \clearfield{urlmonth}%
    \clearfield{urlday}%
  }{}%
}
\AtEveryBibitem{%
  \iffieldundef{doi}{}{%
    \clearfield{url}%
    \clearfield{urldate}%
    \clearfield{urlyear}%
    \clearfield{urlmonth}%
    \clearfield{urlday}%
    \clearfield{eprint}
    \clearfield{eprinttype}
  }%
}
\AtBeginBibliography{\footnotesize}
\def\OJlogo{\vspace{-4pt}$<$Society logo(s) and publication title will appear here.$>$}
\def\seclogo{\vspace{10pt}$<$Society logo(s) and publication title will appear here.$>$}
\def\authorrefmark#1{\ensuremath{^{\textbf{#1}}}}
\input{graphics/bt_commands}

\makeatletter
\def\ps@plain{%
   \def\@oddhead{\vbox{\hsize\textwidth\vspace*{-13pt}\vbox to 29pt{\hsize126pt\vspace*{11.5pt}\OJlogo}\hfill\par\vspace*{-2.75pt}\hbox to \textwidth{\vrule width\textwidth height.3pt depth0pt}}}%
   \let\@evenhead\@oddhead%
   \def\@evenfoot{\vbox to 10pt{\hbox to \textwidth{{\rffont\thepage}\hfill{{\rffont VOLUME\ \@jvol,\ \@pubyear}}}}}%
   \def\@oddfoot{\vbox to 10pt{\hbox to \textwidth{{{\rffont VOLUME\ \@jvol,\ \@pubyear}}\hfill{\rffont\thepage}}}}%
}
\makeatother
\begin{document}

\newif\ifanonymous
\anonymousfalse %

\newcommand{\mosaic}{\textit{\gls{mosaic}}\xspace}

\DeclareRobustCommand{\anontext}[2]{%
  \ifanonymous
    \textcolor{black}{#2}%
  \else
    #1%
  \fi
}

\ifanonymous
\else
    \StopCensoring
\fi

\receiveddate{XX Month, XXXX}
\reviseddate{XX Month, XXXX}
\accepteddate{XX Month, XXXX}
\publisheddate{XX Month, XXXX}
\currentdate{XX Month, XXXX}
\doiinfo{XXXX.2022.1234567}

\markboth{MOSAIC: Modular Supervised Autonomy for Intelligent Coordination of Heterogeneous Robotic Teams}{\xblackout{Oberacker {et al.}}}

\title{MOSAIC: Modular Supervised Autonomy for Intelligent Coordination of Heterogeneous Robotic Teams}

\author{
    {\xblackout{David Oberacker}\authorrefmark{*1,3}\anontext{\orcidlink{0000-0003-2442-4142}}{}}, \and
    {\xblackout{Julia Richter}\authorrefmark{*2}\anontext{\orcidlink{0009-0001-5477-6440}}{}}, \and
    {\xblackout{Philip Arm}}\authorrefmark{2}\anontext{\orcidlink{0000-0001-7666-0177}}{}, \\ \and 
    {\xblackout{Marvin Grosse Besselmann}}\authorrefmark{1}\anontext{\orcidlink{0009-0001-0911-1124}}{}, \and
    {\xblackout{Lennart Puck}}\authorrefmark{1,4}\anontext{\orcidlink{0009-0000-2648-0012}}{}~\xblackout{Member, IEEE}, \and
    {\xblackout{William Talbot}}\authorrefmark{2}\anontext{\orcidlink{0009-0008-9426-3595}}{}, \\ \and
    {\xblackout{Maximilian Schik}}\authorrefmark{1}\anontext{\orcidlink{0009-0006-6272-7370}}{},  \and
    {\xblackout{Sabine Bellmann}}\authorrefmark{1}\anontext{\orcidlink{0009-0008-6085-0353}}{}, \and
    {\xblackout{Tristan Schnell}}\authorrefmark{1}\anontext{\orcidlink{0009-0003-5181-985X}}{}, \and
    {\xblackout{Hendrik Kolvenbach}}\authorrefmark{2}\anontext{\orcidlink{0000-0003-1229-7537}}{}, \\ \and
    {\xblackout{Rüdiger Dillmann}\authorrefmark{1}\anontext{\orcidlink{0000-0002-2049-8219}}{}}~\xblackout{Fellow, IEEE}, \and
    {\xblackout{Marco Hutter}\authorrefmark{2}\anontext{\orcidlink{0000-0002-4285-4990}}{}}~\xblackout{Member, IEEE} \\ \and and
    {\xblackout{Arne Roennau}\authorrefmark{1,3}\anontext{\orcidlink{0000-0002-6090-607X}}{}~\xblackout{Senior Member, IEEE}}
}
\affil{\xblackout{FZI Forschungszentrum für Informatik, Karlsruhe, Germany}}
\affil{\xblackout{Robotic Systems Lab (RSL), ETH Zürich, Zürich, Switzerland}}
\affil{\xblackout{Machine Intelligence and Robotics Lab (MaiRo), Karlsruhe Institute for Technology (KIT), Karlsruhe, Germany}}
\affil{\xblackout{Formerly at FZI Forschungszentrum Informatik, now with the Robotics Section, European Space Agency (ESA), Noordwijk, Netherlands}}
\corresp{
\xblackout{
Corresponding author:
David Oberacker (email: \texttt{oberacker@fzi.de}).
}}
\authornote{
    *Equal contribution\\
    \\
    \xblackout{
    This work was supported by the European Space Agency (ESA) (Ref. 4000141520/23/NL/AT), the Luxembourg National Research Fund (Ref. 18990533), the  German Space Agency (DLR) (Grant Agreements No. 50RA2404) and the Swiss National Science Foundation (SNSF) as part of the projects No.200021E\_229503 and No.227617.
    }
}
\begin{abstract}
\input{sections/00_abstract}
\end{abstract}
\maketitle
\copyrightnotice
\input{sections/01_introduction}
\input{sections/02_related_works}
\input{sections/03_system_architecture}
\input{sections/04_system_description}
\input{sections/05_experiments}
\input{sections/06_evaluation}
\input{sections/07_lessons_learned}
\input{sections/08_conclusion}
\input{sections/xx_acknowledgments}
\printbibliography[title={REFERENCES}] 
\ifanonymous
\else
\input{sections/yy_biography}
\fi
\end{document}

%% file: glossary.tex
\newacronym{ros}{ROS}{Robot Operating System}
\newacronym{ros1}{ROS~1}{Robot Operating System 1}
\newacronym{ros2}{ROS~2}{Robot Operating System 2}

\newacronym[longplural={Points of Interest}]{poi}{POI}{Point of Interest}
\newacronym{lidar}{LIDAR}{Light Detection And Ranging}
\newacronym{qos}{QoS}{Quality Of Service}
\newacronym{dds}{DDS}{Data Distribution Service}
\newacronym{dof}{DoF}{Degree of Freedom}
\newacronym{imu}{IMU}{Inertial Measurement Unit}
\newacronym{ree}{REE}{Rare Earth Elements}
\newacronym{kpi}{KPI}{Key Performance Indicator}
\newacronym{uav}{UAV}{Unmanned Aerial Vehicle}
\newacronym{ugv}{UGV}{Unmanned Ground Vehicle}
\newacronym{xrf}{XRF}{X-ray fluorescence}

\newacronym{mosaic}{MOSAIC}{Modular Supervised Autonomy for Intelligent Coordination of Heterogeneous Robotic Teams}

%% file: graphics/bt_commands.tex
\definecolor{fzi-green}{RGB}{13,114,73}

\definecolor{succeeded}{HTML}{1b5e20}
\definecolor{failed}{HTML}{b71c1c}
\definecolor{running}{HTML}{6A1B9A}%
\definecolor{idle}{HTML}{4E5666}
\definecolor{uninitialized}{HTML}{455a64}
\definecolor{error}{HTML}{CC4700}
\definecolor{shutdown}{HTML}{1a237e}

\definecolor{btbackground}{HTML}{34393C}

\tikzset{
  treenode/.style = {draw=idle, align=center, inner sep=0pt, thick, text centered,
      font=\sffamily, text=white,fill=btbackground,rounded corners,line width=3pt,minimum width=10mm, minimum height=8mm},
  btcondition/.style = {treenode, ellipse, inner sep=4pt},
  btaction/.style = {treenode, rectangle, inner sep=4pt},
  btcapability/.style = {btaction, very thick, dashed},
  btdecorator/.style = {treenode, chamfered rectangle, inner sep=4pt},
  btnode/.style = {treenode, rectangle, inner sep=4pt, text height=0.8em, text width=1.2em},
  round/.style={rounded corners=1.5mm,minimum width=1cm,inner sep=2mm,above right,draw},
  bt-idle/.style={draw=idle, thick},
  bt-uninitialized/.style={draw=uninitialized, thick},
  bt-succeeded/.style={draw=succeeded, thick},
  bt-failed/.style={draw=failed, thick},
  bt-running/.style={draw=running, thick},
  bt-error/.style={draw=error, thick},
  bt-shutdown/.style={draw=shutdown, thick}
}

\tikzset{fontscale/.style = {font=\relsize{#1}}}

%% file: sections/00_abstract.tex
Mobile robots have become indispensable for exploring hostile environments, such as in space or disaster relief scenarios, but often remain limited to teleoperation by a human operator.
This restricts the deployment scale and requires near-continuous low-latency communication between the operator and the robot.
We present~\textit{\acrshort{mosaic}}: a scalable autonomy framework for multi-robot scientific exploration using a unified mission abstraction based on~\acrfullpl{poi} and multiple layers of autonomy, enabling supervision by a single operator.
The framework dynamically allocates exploration and measurement tasks based on each robot's capabilities, leveraging team-level redundancy and specialization to enable continuous operation.
We validated the framework in a space-analog field experiment emulating a lunar prospecting scenario, involving a heterogeneous team of five robots and a single operator.
Despite the complete failure of one robot during the mission, the team completed 82.3\% of assigned tasks at an Autonomy Ratio of 86\%, while the operator workload remained at only 78.2\%.
These results demonstrate that the proposed framework enables robust, multi-robot scientific exploration with limited operator intervention.
We further derive practical lessons learned in robot interoperability, networking architecture, team composition, and operator workload management to inform future multi-robot exploration missions.

%% file: sections/01_introduction.tex
\begin{figure*}[t]
    \centering
    \includegraphics[width=\textwidth]{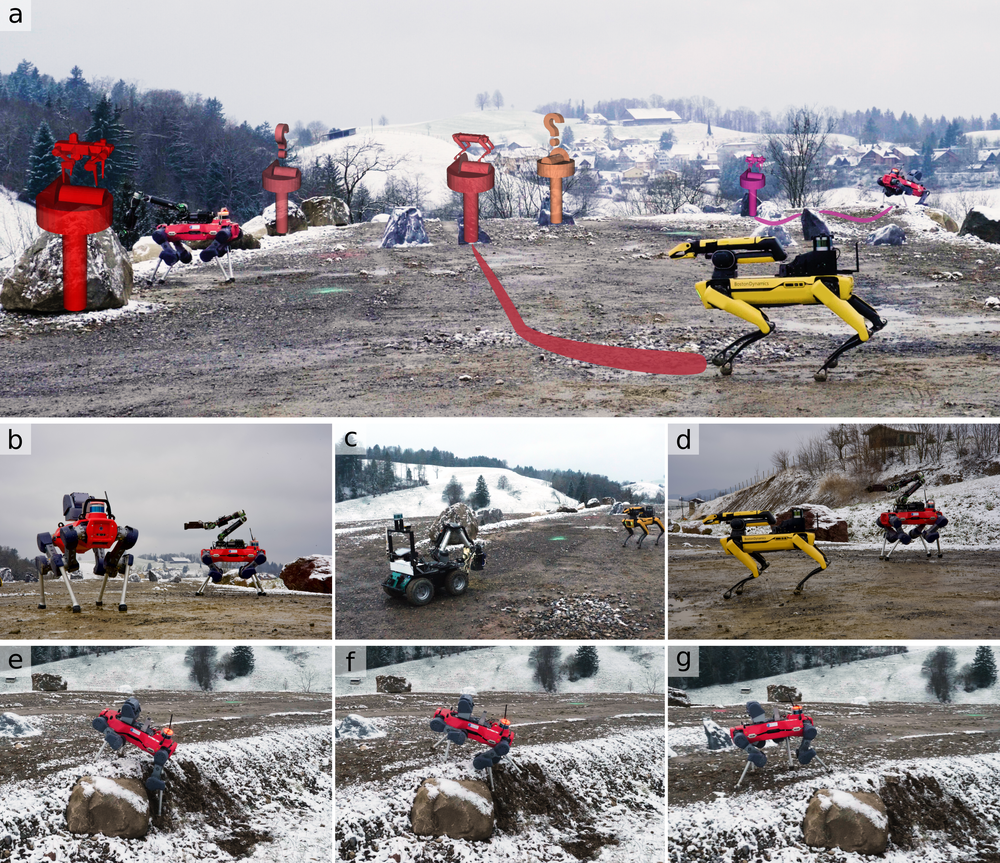}
    \caption{
        Field experiments demonstrating multi-robot operation and locomotion capabilities in unstructured outdoor terrain.
        (a) Representative example of the test site with \gls{poi} visualization overlaid on the map (\textit{?}: unclaimed, robot icon: currently executing, lines: planned path)
        (b-d) Representative examples of the test site with multiple platforms operating concurrently.
        (e-g) \anontext{Dilly}{ANYmal~2} traversing a steep, loose slope, illustrating climbing behavior and stability during challenging terrain interaction.
    }
    \label{fig:mission_insights}
\end{figure*}
\section{INTRODUCTION}
\label{sec:INTRODUCTION}
\lettrine{M}{obile} robots have long been employed in environments that are dangerous or inaccessible to humans.
Applications such as extraterrestrial exploration, search and rescue, or hazard containment (e.g., nuclear accidents or bomb defusals) are prime examples where robots have been established as a fundamental tool.
These robots currently operate primarily as teleoperated systems with limited assistance functions~\cite{seo_how_2021, rea_still_2022, orekhov_inspiring_2023, schmaus_toward_2023}, providing support during complex tasks, such as ensuring collision-free navigation.
As a result, they require continuous supervision and input from at least one skilled operator to ensure correct behavior, leading to low operational efficiency~\cite{rea_still_2022}.
\par
Historically, many space missions have favored single-robot designs, as they minimize operational overhead and reduce system integration between multiple platforms.
In practice, however, such designs require the integration of diverse sensing, actuation, and capabilities into a single platform, substantially increasing hardware and software complexity.
This is evident in the architecture of the latest Mars rovers, Curiosity~\cite{cook_curiosity_2012} and Perseverance~\cite{sun_overview_2023}, which integrate numerous subsystems.
In such single-agent setups, any critical failure can end the entire mission. 
This is widely discussed in \cite{prorok_beyond_2021} and was also evident during the \emph{Mars Exploration Rover Mission}, where the Spirit rover became stuck on a slope in an orientation that proved fatal during the Martian winter, ending the mission early \cite{nasa__jpl_mars_nodate}.
\par
To address these issues, researchers have begun to shift towards robotic team-based solutions.
Challenges like the DARPA Subterranean Challenge~\cite{orekhov_inspiring_2023} or ESA Space Resources Challenge~\cite{european_space_agency_challenge_nodate} have shown how the multi-robot paradigm has measurable advantages in terms of robustness and operational efficiency compared to single robot approaches~\cite{tranzatto_cerberus_2022,kottege_heterogeneous_2025,arm_scientific_2023,schnell_efficient_2023, van_der_meer_realms_2023}.
The inherent \textbf{robustness} of multi-robot teams stems from two forms of redundancy:
First, by allowing for fault tolerance at the robot level, tasks can be performed by multiple team members.
Secondly, in a heterogeneous team, different systems capable of performing the same task can be introduced, enabling greater robustness and flexibility in task allocation.
In terms of \textbf{operational efficiency}, team approaches can leverage both task parallelism and specialization to increase throughput.
Instead of a single generalized but suboptimal system, individual robots can be optimized for specific task subsets.
While reducing the system's redundancy, such specialization enables the use of higher-quality hardware or efficiency gains through more tailored kinematics.
\par
Beyond team composition, the choice of locomotion modality remains an open question.
While wheeled rovers have been the established platform for planetary exploration, advances in legged locomotion have made such platforms increasingly competitive in unstructured outdoor terrain.
Legged systems are particularly attractive for accessing scientifically interesting but locomotion-challenging regions such as steep crater walls or lava tubes, which exceed the capabilities of conventional rovers~\cite{kolvenbach2025lunarleaper}.
This motivation has driven a sustained line of prototype work on legged platforms for planetary exploration~\cite{bartsch2010spaceclimber, seidel2020using, olsen2025olympus}.
Combined with the team-based paradigm, heterogeneous platforms with mixed locomotion modalities can leverage the complementary strengths of both: legged platforms for terrain access, wheeled platforms for endurance and payload capacity.
\par
With large teams of primarily manually controlled robots, the model of continuous operator supervision does not scale well: either the number of human operators must increase, or robots experience longer idle times due to delayed input, both of which negatively impact overall mission efficiency.
To offset efficiency losses and manpower demands, higher levels of autonomy are required to allow a single operator to effectively supervise and coordinate a multi-robot team.
We therefore design the MOSAIC framework to enable a single operator to supervise and control a team of robots by dynamically adjusting the level of autonomy of individual robots or the full team.
Realizing this mixed autonomy operation of a multi-robot team introduces several challenges:
(1) Robots must operate in a shared environment, requiring, for example, that each robot filters out other team members from its local occupancy map to avoid treating them as static obstacles;
(2) Exploration and scientific measurement tasks need to be allocated across the team according to each robot's capabilities, while optimizing mission parameters such as time or energy, and
(3) Operators must retain the ability to intervene to ensure safe operation.
\par
To demonstrate the practical viability of~\mosaic, we conducted exploration and analysis missions with a heterogeneous team of robots, coordinated by a single operator, under real-world conditions.
For quantitative evaluation of the mission outcome, we applied a subset of the~\gls{kpi} framework proposed in~\cite{richter_practical_2026}, which was specifically designed for multi-robot exploration missions.
We selected~\glspl{kpi} covering three dimensions, efficiency, robustness, and scientific precision, evaluated under realistic deployment conditions.
\par
Although multi-robot systems have been demonstrated in diverse environments, mission-level autonomy, where robots autonomously generate, allocate, and execute mission objectives without operator input, has yet to be fully achieved.
In this work, we take an important step toward this goal with the following contributions:
\begin{enumerate}
    \item We introduce and implement~\mosaic, a scalable autonomy framework that enables parallelized scientific exploration of an area by a team of robots under the supervision of a single operator at runtime.
    This parallelized exploration is enabled through a flexible objective-distribution system and an intuitive user interface that allows shaping the system's in-mission behavior.
    \item We propose a unified representation of mission objectives as~\glspl{poi}, which are converted to robot-specific tasks for execution on an assigned system.
    \item We evaluate the proposed system in a space-analog field experiment (\cref{fig:mission_insights}) to assess its effectiveness under real-world conditions.
    \item We provide a comprehensive list of lessons learned to outline the challenges and insights of the system and field tests, with particular emphasis on working with the \gls{ros2} framework.
\end{enumerate}
\par
The remainder of the paper is structured as follows.
\Cref{sec:RELATED_WORKS} discusses related work.
The architecture of~\mosaic is introduced in~\cref{sec: SYSTEM ARCHITECTURE}, with a detailed description of all robots and components in~\cref{sec: SYSTEM DESCRIPTION}.
\Cref{sec:EXPERIMENTS} describes the mission setup, followed by the evaluation in~\cref{sec:RESULTS}.
The lessons learned are presented in~\cref{sec: LESSONS LEARNED}.
\Cref{sec: CONCLUSION} summarizes the work and presents future directions.

%% file: sections/02_related_works.tex
\section{RELATED WORKS}
\label{sec:RELATED_WORKS}
A variety of heterogeneous multi-robotic teams are employed in different scenarios, often motivated by large-scale robotics challenges. 
One significant enabler was the DARPA Subterranean (SubT) Challenge~\cite{orekhov_inspiring_2023}, in which teams explored unknown caves, detected artifacts, and reported their locations under severely constrained communication. 
Beyond subterranean robotics, multi-robot systems have also been evaluated in search and rescue \cite{queralta2020collaborative} and planetary analog missions. The ESA Space Resources Challenge (SRC)~\cite{european_space_agency_challenge_nodate} simulated a lunar surface prospecting scenario that required the autonomous detection of boulders and \gls{ree} oxide patches.
\par
Building on these deployments, we review prior work by examining how different systems address key challenges that arise during multi-robot exploration missions, with a particular emphasis on task management during execution, autonomy and human interaction, and scalability in heterogeneous teams.
We focus on methods that have been validated in large-scale field experiments rather than purely theoretical studies, as these demonstrate practical feasibility under real-world conditions. %
\paragraph{Online Task Management} 
In this section, we follow the terminology of the cited works, in which the term task is used broadly for both mission-level goals and the robot-specific actions required to achieve them. In MOSAIC, we explicitly distinguish these two concepts (see Sec.~\ref{subsubsed: MISSION OBJECTIVES}).
While coarse prior information is often available before the deployment of an exploration mission, it is typically insufficient to anticipate all task-relevant findings that emerge during execution.
As robots explore and encounter obstacles, revisit previously surveyed areas, or identify new points of scientific interest, the mission's task landscape evolves continuously.
Robots operating at higher levels of autonomy must therefore be capable of reacting to unforeseen situations and incorporating newly discovered information into their ongoing task management.
This dynamic emergence of mission-relevant information necessitates flexible online task generation, sequencing, and allocation across the multi-robot team.
\par
The types of tasks assigned to robots vary with the mission objectives.
Most demonstrations focus on navigation tasks, in which the robot autonomously navigates to a waypoint while avoiding obstacles or hazardous areas.
Such tasks were central not only in SubT \cite{tranzatto_cerberus_2022, biggie_flexible_2023, cao_autonomous_2022} but also in planetary-analog missions including Skylight Exploration~\cite{dominguez_cooperative_2025},  LUNARES \cite{cordes_lunares_2011}, AMADEE-24 \cite{jusner_mission_2024}, and ARCHES \cite{schuster_arches_2020}.
Beyond navigation, science-focused campaigns also incorporate tasks such as spectral rock analysis \cite{schuster_arches_2020, van_der_meer_realms_2023, arm_scientific_2023} or sample localization and retrieval \cite{cordes_lunares_2011, sonsalla_field_2017}, which require accurate target identification, stable positioning relative to the environment, and, in some cases, physical interaction with specific objects.
\par
The strategies for distributing and prioritizing exploration tasks vary substantially across systems.
In SubT, several teams employed autonomous exploration architectures in which robots autonomously generate waypoints from frontier regions without operator input \cite{tranzatto_cerberus_2022, biggie_flexible_2023, cao_autonomous_2022}. 
Typically, a global planner guides each agent toward frontiers expected to maximize information gain~\cite{tranzatto_cerberus_2022, biggie_flexible_2023}.
These planners can be implemented in a robot-centric manner, where each robot independently optimizes its own exploration progress \cite{tranzatto_cerberus_2022}, or extended to allow plan exchange and adaptation whenever communication becomes available \cite{biggie_flexible_2023}.
Such approaches are particularly effective in cave environments, where tunnel-like geometries strongly constrain feasible paths and naturally limit redundant exploration.
However, in open environments with less constrained geometry, the lack of explicit coordination in robot-centric frontier planners can lead to overlapping exploration efforts and reduced team-level efficiency.
\par
Autonomous robot task planners face additional complexity when missions extend beyond exploration to include scientific objectives, particularly when tasks involve time-consuming, precision-critical measurements or rely on specialized instruments available only on specific agents.
Some systems address this complexity through predefined autonomous task sequences, most notably in structured sample-return experiments \cite{sonsalla_field_2017, cordes_lunares_2011}.
More commonly, task assignment is carried out online through human-in-the-loop task planning, with operators continuously analyzing mission data to plan and issue navigation waypoints, define science targets, or trigger measurement actions during execution~\cite{schuster_arches_2020, jusner_mission_2024, arm_scientific_2023}.
While this approach supports efficient exploration, it places additional demands on the operator, which become more pronounced as team size increases and can reduce overall scalability.
One fully autonomous solution for robot-specific task assignment is provided by task-bidding frameworks \cite{hudson_heterogeneous_2022, kottege_heterogeneous_2025, schnell_efficient_2023}, which explicitly encode task requirements and robot capabilities, ensuring that scientific tasks are assigned only to agents equipped with the required sensors or instruments.
This autonomous solution reduces the need for continuous operator involvement and facilitates task reallocation when team composition changes due to failures or additional deployments during mission runtime.
\par
Task-bidding frameworks represent one approach to autonomous task allocation, alongside optimization-based methods and learning-based techniques such as multi-agent reinforcement learning~\cite{chakraa2023optimization}.
However, optimization-based methods become computationally infeasible as the number of robots and targets grows, and struggle to accommodate dynamic changes such as newly discovered targets during mission execution~\cite{chakraa2023optimization}. Learning-based methods address runtime scalability by shifting computational demand to training time, but training times and memory requirements scale rapidly with team size and number of targets. Furthermore, fixed observation sizes require retraining when team composition changes~\cite{rubio2026collaborative}. Both paradigms, therefore, present practical barriers for our intended scenario.
More recently, large language models have been proposed as a coordination layer for multi-robot teams~\cite{huang2025compositional}, enabling natural-language mission specification.
However, such approaches rely on continuous connectivity to external inference infrastructure, which is incompatible with the communication-constrained deployments targeted by MOSAIC.

\paragraph{Autonomy Modes and Human-in-the-Loop Operation}
While advances in autonomy enable robots to plan, manage, and execute a wide range of tasks independently, real-world deployments inevitably expose situations that cannot be reliably anticipated or estimated during mission execution. 
Practical systems must therefore still support human involvement during execution, allowing operators to intervene, adapt mission priorities, or recover from failures when autonomous decision-making proves insufficient.
\par
The task management strategies discussed above thus operate within a set of autonomy modes that recur across prior deployments. 
The highest autonomy mode is \emph{mission-level autonomy}, in which robots autonomously generate tasks, allocate them across the team, and execute them without operator input. 
Only systems that support autonomous task discovery fall into this category~\cite{schnell_efficient_2023, biggie_flexible_2023, hudson_heterogeneous_2022}. 
A second mode is \emph{task-level autonomy}, where a human operator specifies tasks that robots then execute autonomously.
While this capability is present in many of the aforementioned systems, it is particularly emphasized in~\cite{schuster_arches_2020} and~\cite{sonsalla_field_2017}. 
The lowest autonomy mode is \emph{teleoperation} or \emph{driver-level autonomy}, which overrides task-level or mission-level autonomy and is typically used for recovery or fine-grained intervention. 
\par
In~\mosaic, we adopt this three-level autonomy model and aim to maximize the duration of operation in the highest autonomy mode.
\paragraph{Team Composition and Scalability}
Scalability is a central challenge in multi-robot autonomy, as increasing team size,  but also increasing heterogeneity, places growing demands on coordination, communication, and supervision.  
While robotic swarm systems have demonstrated teams of dozens or even hundreds of agents, particularly in homogeneous \gls{uav} deployments \cite{preiss_crazyswarm_2017}, these systems typically rely on uniform platforms and sensing payloads, which simplifies these challenges.
\par
Hardware-heterogeneous robotic teams, on the other hand, are substantially smaller. 
In SubT, the largest team comprised up to 12 robots \cite{agha_nebula_2021}. 
Such teams commonly combine legged robots, which provide access to difficult terrain through advanced locomotion capabilities \cite{biggie_flexible_2023, agha_nebula_2021, tranzatto_cerberus_2022}, wheeled platforms that offer higher energy efficiency and longer endurance \cite{cao_autonomous_2022} or serve as mobile communication relays \cite{biggie_flexible_2023, cao_autonomous_2022}, and aerial vehicles used for rapid exploration and situational awareness \cite{agha_nebula_2021, tranzatto_cerberus_2022}. 
Despite this platform diversity, sensing payloads in SubT were largely homogeneous, primarily comprised of LiDARs, cameras, and IMUs, which simplified system integration but limited sensing specialization.
\par
Space-analog missions introduce additional challenges that limit scalability due to the need for specialized scientific instruments, which increase integration effort and complicate autonomous task allocation \cite{arm_scientific_2023, schnell_efficient_2023}. 
As a result, deployments rely on smaller teams of two to three robots with heterogeneity in both platform and sensor configuration, such as drones for scouting and mapping \cite{schuster_arches_2020}, rovers equipped with spectrometers or manipulators \cite{schuster_arches_2020, cordes_lunares_2011}, and legged robots for rapid or terrain-challenging scouting \cite{arm_scientific_2023}.
\par
Scaling such specialized teams remains difficult.  
As the number of robots and task types increases, operator workload grows accordingly, particularly when scientific tasks require detailed supervision, validation, or interpretation of results \cite{schuster_arches_2020, arm_scientific_2023}.  
In addition, the task management must remain robust in the face of intermittent communication losses and partial system failures, such as individual agents becoming unavailable during mission execution.
This trade-off between team size, autonomy, and operator effort continues to limit the practical scalability of heterogeneous, science-driven multi-robot systems.

%% file: sections/03_system_architecture.tex
\section{AUTONOMY CONCEPT \& ARCHITECTURE}
\label{sec: SYSTEM ARCHITECTURE}
This section presents the system architecture underlying~\mosaic, our scalable autonomy framework for heterogeneous, science-driven multi-robot missions.
We first introduce our system's core autonomy concepts, including the representation of mission objectives, multiple levels of autonomy, and mission-level robot roles.
Together, these concepts define how mission objectives are formulated, how autonomy is shared between human operators and robots, and how heterogeneous platforms can contribute to mission-relevant outcomes.
Building on this conceptual foundation, we then describe the autonomy architecture that realizes these ideas in practice.
\subsection{Autonomy Concept}
\label{subsec: AUTONOMY CONCEPT}
The level of autonomy defines the degree of involvement of a human in a robot's decision-making process during runtime~\cite{richardson_systematic_2025}.
This ranges from a human determining all system actions (full teleoperation) to systems with no human involvement (full mission-level autonomy).
Realizing effective \textbf{autonomy management} while accounting for applicability to smaller-to-medium-sized teams requires a system built on unified interfaces that allow heterogeneity at lower levels.
\textbf{Scalable autonomy} in this context is defined as a system in which robots operate at variable levels of autonomy, which can be scaled up (more autonomous) or down (more human involvement) at any time.
The following concept outlines how we use scalable autonomy to manage a team of robots that fulfill different roles within the team.
We also detail how mission objectives are defined and distributed to the robots on a system level.
\subsubsection{Mission Objectives}
\label{subsubsed: MISSION OBJECTIVES}
Scalable autonomy in science-driven exploration missions requires a clear and consistent representation of what the robot team is expected to achieve.
We therefore define \textbf{mission objectives} as the primary unit of mission progress, capturing outcomes such as exploring an area, collecting a sample, or analyzing a rock.
In the target mission class, every such objective is tied to a specific location in space.
We exploit this property by representing each objective as a \gls{poi} that pairs the spatial location with a type specifying the required outcome.
While this deviates from the common usage of the term~\glspl{poi} in mapping, it underlines this assumption and provides a unified representation of objective and location.

We further distinguish a mission objective from the tasks required to fulfill it. A task is a robot-specific, executable action sequence, for instance, planning and following a path, operating a manipulator, or triggering a scientific instrument, that contributes to the completion of an assigned objective. 
The mapping from objectives to tasks depends on each robot's capabilities and is resolved at the task level (Sec.~\ref{subsubsed: AUTONOMY LEVELS}).
This abstraction allows us to decouple what needs to be achieved (mission outcome) from how it is achieved (robot-specific execution), ensuring that objectives remain consistent and comparable across the robot team.

\subsubsection{Autonomy Levels}
\label{subsubsed: AUTONOMY LEVELS}
As defined in~\cref{subsec: AUTONOMY CONCEPT}, systems can operate at variable levels of autonomy.
In ideal conditions, robots would complete all mission objectives fully autonomously; however, real-world missions are highly unpredictable and often require varying degrees of operator intervention.
Building on the system proposed in~\cite{schnell_efficient_2023},~MOSAIC allows the operator to adjust the autonomy level of individual robots or the full team and intervene at the appropriate level of abstraction.
To maximize team efficiency, robots are designed to operate at the highest feasible level of autonomy. %
\paragraph{Mission Level}
At this level, the~\mosaic operates on mission objectives that contribute directly to overall mission progress.
Objectives are associated with locations in the shared mission map, requiring consistent co-localization across the robot team.
During operation on mission-level autonomy, robots independently select and claim objectives based on their assessment of the current mission state, while exchanging future plans to avoid conflicts.
Objectives may be generated by the operator or dynamically created by robots during execution.
Operator interaction at this level focuses on supervising mission progress and prioritizing objectives, with any intervention affecting the entire robot team.
\paragraph{Task Level}
The task level describes the autonomous behavior of individual robots during task execution, i.e., the execution of the action sequences (defined in Sec. III.A.1) that realize an assigned mission objective on a specific platform.
It executes the actions required to complete an assigned objective, including navigation, manipulation, and scientific data acquisition, as well as maintenance behaviors such as battery management.
This level integrates the robot's core autonomy components and executes task-specific behaviors.
Operator interventions at this level directly modify the robot's behavior, for instance, by disabling certain actions or commanding motion to a specific pose.
\paragraph{Driver Level}
The driver level provides access to the robot's hardware, enabling the execution of fine-grained control commands directly on actuators or via low-level abstractions (such as locomotion policies).
At this level, the robot does not pursue any task or objective autonomously; instead, control flow is determined entirely by operator input. Assistive functions, such as preventing the robot from walking into an obstacle, may still be active in the background to reduce operational risk, but they only constrain commanded motion rather than initiate behavior.
Intervention at the driver level offers maximum control flexibility but requires continuous operator input.
This distinguishes itself from task-level commands as these are executable control commands, which are not specific to the mission or the current context.
\subsubsection{Robot Roles}
In addition to the three levels of autonomy, we propose distinct mission-level skill sets for robots, serving as roles with unique capabilities.
A role is defined by a specialized set of mission-relevant capabilities that a robot must possess to perform objectives integral to the mission.
Defining multiple roles in a multi-robot team simplifies the online task management by providing a clear list of capabilities for each robot.
Well-defined roles can also account for unknown environments by providing a diverse skill set to the overall team.
Each role should thereby be realized by at least two distinct robots, where robots may also implement multiple roles at once.
This role differentiation enables both specialization and redundancy on the mission level, since each role is performed by multiple robots to ensure continuous coverage of all required capabilities.
\subsection{Autonomy Architecture}
\label{subsec: SOFTWARE ARCHITECTURE}
We realize the autonomy concepts introduced above in a decentralized manner, requiring a software architecture that balances decentralized task execution with centralized mission coordination.
In particular,~\mosaic must maintain a consistent global mission state, provide effective operator interfaces, and allow individual robots to operate autonomously and robustly in the field.

\begin{figure*}[t]
    \centering
    \includegraphics[height=0.5\textheight]{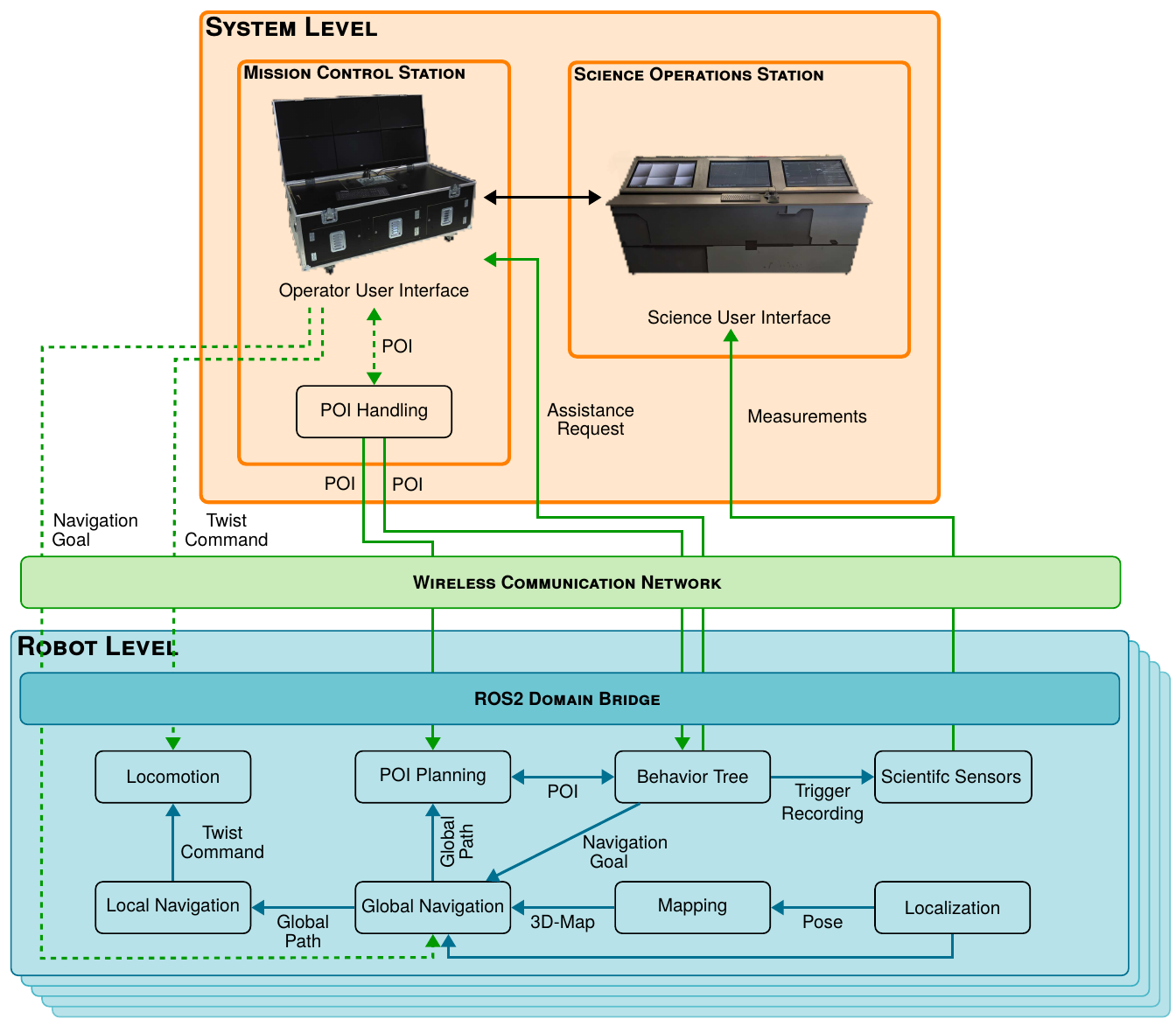}
    \caption{
        \mosaic software architecture overview of both individual robots and the overall control system.
        The~{\color{orange} system-level components} exist once in the system,~{\color{MidnightBlue} robot-specific components} exist on every robot in the team.
        Communication is achieved on a per-robot {\color{MidnightBlue}unique~\gls{ros2} domain}.
        Shared communication happens on the~{\color{OliveGreen} shared~\gls{ros2} domain}.
        }
    \label{fig: system architecture}
\end{figure*}

We therefore organize the autonomy architecture into two complementary layers.
At the \textbf{system level}, shared services manage mission objectives, maintain global consistency, and interface with human operators through a centralized \textit{mission control}.
The system level is primarily concerned with operations on the \textit{mission level} on our autonomy scale, but also provides the operator with the interface for interaction with the \textit{task} and \textit{driver level}. 
At the \textbf{robot level}, local autonomy components are responsible for task planning, navigation, and execution on each individual platform.
This separation enables scalable supervision of heterogeneous robot teams while preserving robot-level independence.
An overview of the resulting architecture and the interaction between system-level and robot-level components is shown in~\Cref{fig: system architecture}.
\subsubsection{System-Level Components}
The system level (highlighted {\color{orange} orange} and {\color{OliveGreen} green} in~\Cref{fig: system architecture}) provides the shared infrastructure required for mission coordination, operator interaction, and global state management.

\begin{figure*}
    \centering
    \subfloat[
        Mission control station with the following UI elements:
        \textit{Top row:} Spot control interface with its camera feed showing \anontext{Donkey}{ANYmal~3} performing a ground measurement (left), RViz view for placing and interacting with various POI types (middle), and thermal image of the testing area captured by \anontext{Dilly}{ANYmal~2} (right) with its control interface.
        \textit{Bottom row:} behavior tree state (left), teleoperation interface (middle), and the onboard camera of \anontext{Donkey}{ANYmal~3} showing the robotic arm during a ground measurement (right).
        \label{fig:base_station_ui:base_station}
    ]{
        \includegraphics[width=.9\textwidth]{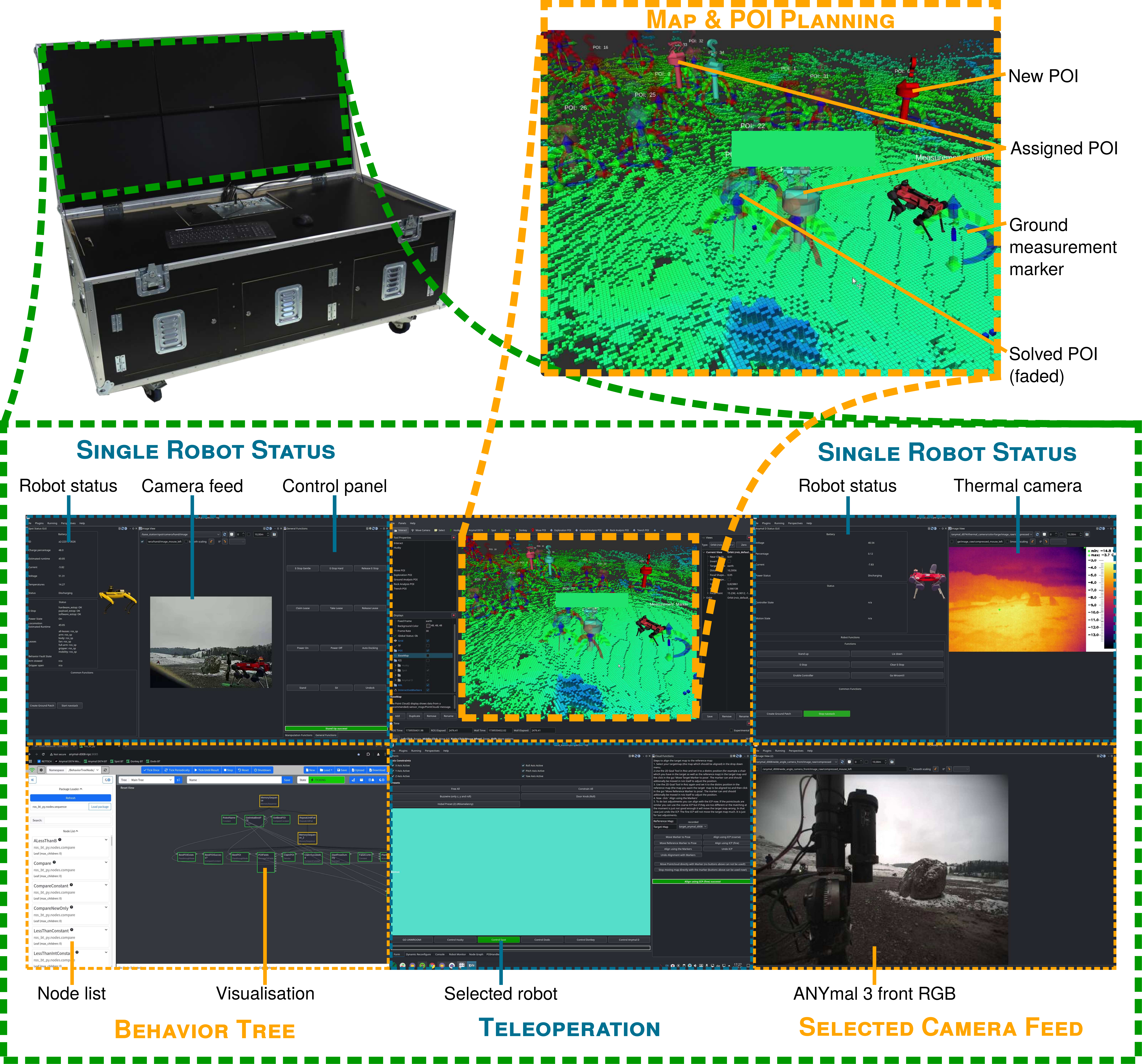}
    }\\
    \subfloat[
        Science operations station with the following UI elements:
        \textit{Left:} Collected data from a measurement synced from the robot.
        \textit{Right:} Refinement of scientific measurement poses utilizing images from the robot's arm cameras.
        \label{fig:base_station_ui:science_station}
    ]{
        \includegraphics[width=.9\textwidth]{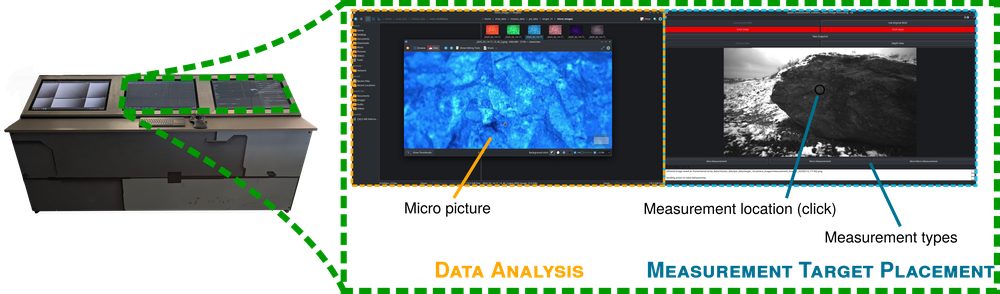}
    }
    \caption{
        Custom-built mission control (a) and touchscreen science operations station (b) with key interface elements highlighted.
    }
    \label{fig:base_station_ui}
\end{figure*}

The \textit{Mission Control} serves as the primary interface between the human operators and the robot team.
It consists of two workstations: an \textit{Mission Control Station} and a \textit{Science Operations Station}.
The mission control station provides all interfaces for monitoring and controlling the robot team, including mission supervision, objective management, teleoperation, and status visualization.
The science operations station supports researcher-in-the-loop workflows by enabling domain experts to inspect incoming data, assess measurement quality, and provide feedback that influences ongoing mission execution.
As illustrated in~\Cref{fig:base_station_ui}, the operator station employs a multi-screen layout that combines a shared environment map with real-time robot poses, planned navigation paths, and all active \glspl{poi}, supporting effective mission supervision and control.
\par
Mission objectives, represented as \glspl{poi} as introduced earlier, are managed at the system level through a centralized \textit{\gls{poi} Handling} component.
Each~\gls{poi} is defined by a set of attributes that describe its location, type, assignment state, and planning-related metadata, as summarized in~\Cref{fig: poi}.
A POI has a unique ID and a \textit{pose} in the shared \textit{earth} frame, and its \textit{type} specifies the required objective.
The \textit{robot} field indicates which robot is currently assigned, while the fields \textit{mission\_value}, \textit{robot\_attempts}, and \textit{robot\_utilities} are used for planning and selection.
A POI is created in the \textit{active} state and is set to inactive once a robot completes the objective or the operator removes it.
They are managed at the system level, specifically on the mission control station, to maintain consistency in the mission state across robots.
\begin{algorithm}
    \caption{Definition of the POI data structure.}
    \label{fig: poi}
    \begin{algorithmic}%
        \Struct{POI}
          \State $id$ : \textsc{Int}
          \State $pose$ : \textsc{Pose}
          \State $type$ : \textsc{String}
          \State $robot$ : \textsc{String}
          \State $active$ : \textsc{Boolean}
          \State $mission\_value$ : \textsc{Double}
          \State $robot\_attempts$ : \textsc{[RobotAttempt]}
          \State $robot\_utilities$ : \textsc{[RobotUtility]}
        \EndStruct
        \Struct{RobotAttempt}
            \State $stamp$ : \textsc{Timestamp}
            \State $robot$ : \textsc{String}
            \State $success$ : \textsc{Boolean}
        \EndStruct
        \Struct{RobotUtility}
            \State $robot$ : \textsc{String}
            \State $utility\_value$ : \textsc{Double}
        \EndStruct
    \end{algorithmic}
\end{algorithm}
Lastly, a \textit{Wireless Communication Network} connects all robots to mission control, enabling the exchange of mission state, objective information, and operator commands.
This network forms the backbone of the~\mosaic framework, allowing robots to operate independently while remaining synchronized through shared system-level services.
Together, the mission control, POI handling, and communication infrastructure establish the global context of our scalable multi-robot autonomy system.
\subsubsection{Robot-Level Components}
While the system level manages coordination and shared mission state, the robot level (highlighted {\color{MidnightBlue} blue} in~\Cref{fig: system architecture}) is responsible for executing individual tasks.

The \textit{\gls{poi} Planning} component runs locally on each robot and selects upcoming objectives based on the current mission state.
From the set of active \glspl{poi} provided by mission control, the robot computes a utility for each task that reflects its capabilities and the intentions of other robots (obtained from each robot's broadcast utilities), constituting a lightweight form of anticipatory coordination~\cite{skulimowski2025cooperation}.
Since the robots differ in locomotion, manipulation, and sensing abilities, the rewards associated with each POI are calculated robot-specifically.
Details on the reward calculation (\textit{utilities}) can be found in the next~\cref{subsubsec:poi_planning}.
The planner is invoked from a behavior tree and rerun whenever a new objective is needed.

High-level decision-making on each robot is implemented using a \textit{Behavior Tree}.
The tree coordinates objective selection, task execution, and teleoperation overrides within a unified control structure.
This structure provides a predictable and transparent control flow, which is particularly important in multi-robot missions where individual robots may be temporarily disconnected or interrupted by an operator.
Task execution is realized through dedicated subtrees selected at runtime based on the robot and \gls{poi} type.
Together with the \gls{poi} planner, the behavior tree forms the core of the robot's autonomy loop, linking mission-level objective allocation to local task execution.

Additionally, each robot has components providing the foundational capabilities necessary to navigate the environment and perform assigned objectives.
\textit{Global Navigation} generates a coarse, long-range path toward the objective goal using a static snapshot of the environment, enabling efficient planning over large distances.
\textit{Local Navigation} refines this path in real time, reacting to obstacles, terrain irregularities, and other robots to ensure safe and feasible motion.
The \textit{Mapping} component maintains a geometric representation of the environment that supports both planning stages and is continuously updated as the robot explores.
\textit{Localization} estimates the robot's pose within the shared reference frame, allowing objectives and paths to remain spatially consistent across the team.
Finally, the \textit{Locomotion} components translate navigation commands into platform-specific motions, enabling reliable traversal of diverse terrain despite differences in robot morphology.
To maintain the autonomy framework's independence from robot hardware, each module exposes standardized interfaces, such as pose goals for navigation or twist commands for teleoperation, regardless of the underlying robot platform.

This abstraction is essential for heterogeneous missions, where different robots may rely on different navigation stacks or locomotion controllers.
By standardizing how~\mosaic autonomy components interact with these modules, the system can scale across multiple robot types with minimal adjustments.
A more detailed description of these modules is given in the following chapter, where our individual robot platforms are discussed.

%% file: sections/04_system_description.tex
\section{SYSTEM IMPLEMENTATION}
\label{sec: SYSTEM DESCRIPTION}
This section describes the technical realization of~\mosaic, our heterogeneous multi-robot team setup, described in~\Cref{sec: SYSTEM ARCHITECTURE}.
First, we give a detailed overview of the five robots integrated into the system.
Secondly, we describe the practical realization of the~\mosaic autonomy framework introduced in~\cref{subsec: AUTONOMY CONCEPT}.
Next, the shared mapping and how we navigate using this map information are presented.
We also outline the operator's responsibility, including in the gathering of scientific measurements.
Lastly, we present the detailed communication architecture.
\par
All software components of~\mosaic are built on the~\gls{ros} middleware, with mission-level components running on \acrshort{ros2} Humble~\cite{macenski_robot_2022}, and robot-level components as well as lower-level drivers implemented in a mix of \gls{ros1} Noetic~\cite{morgan_quigley_ros_2009} and \gls{ros2} Humble.
Communication between components is handled through~\gls{ros} messages, services, and actions, depending on the required interaction.
\par
As introduced in~\cref{subsec: AUTONOMY CONCEPT}, we separate the robots into two roles, scouts and scientists.
Scouts are responsible for exploration and image-level analysis.
They are designed to locomote faster and more agilely, but carry no specialized scientific instruments.
Scientists, on the other hand, specialize in acquiring scientific data from the mission area.
They are equipped with dedicated instruments for analysis objectives, which increases payload weight and reduces mobility compared to scouts.
Consequently, to effectively use their specialized instruments, scientists typically follow scouts to perform analysis in already-explored areas.
\subsection{Scouts}
\label{subsec: SCOUTS}
Scouts are tasked with exploration and initial information gathering and are equipped with \gls{lidar} and RGB-D cameras as their main instruments.
This information is processed to automatically generate \glspl{poi} at notable locations for the scientists to investigate further.
We employed three robots as scouts, two heterogeneous ANYbotics ANYmal~D~\cite{anybotics_ag_anymal_2025} and one Boston Dynamics Spot~\cite{boston_dynamics_inc_spot_2025}.
\begin{figure*}[t]
    \includegraphics[width=\linewidth]{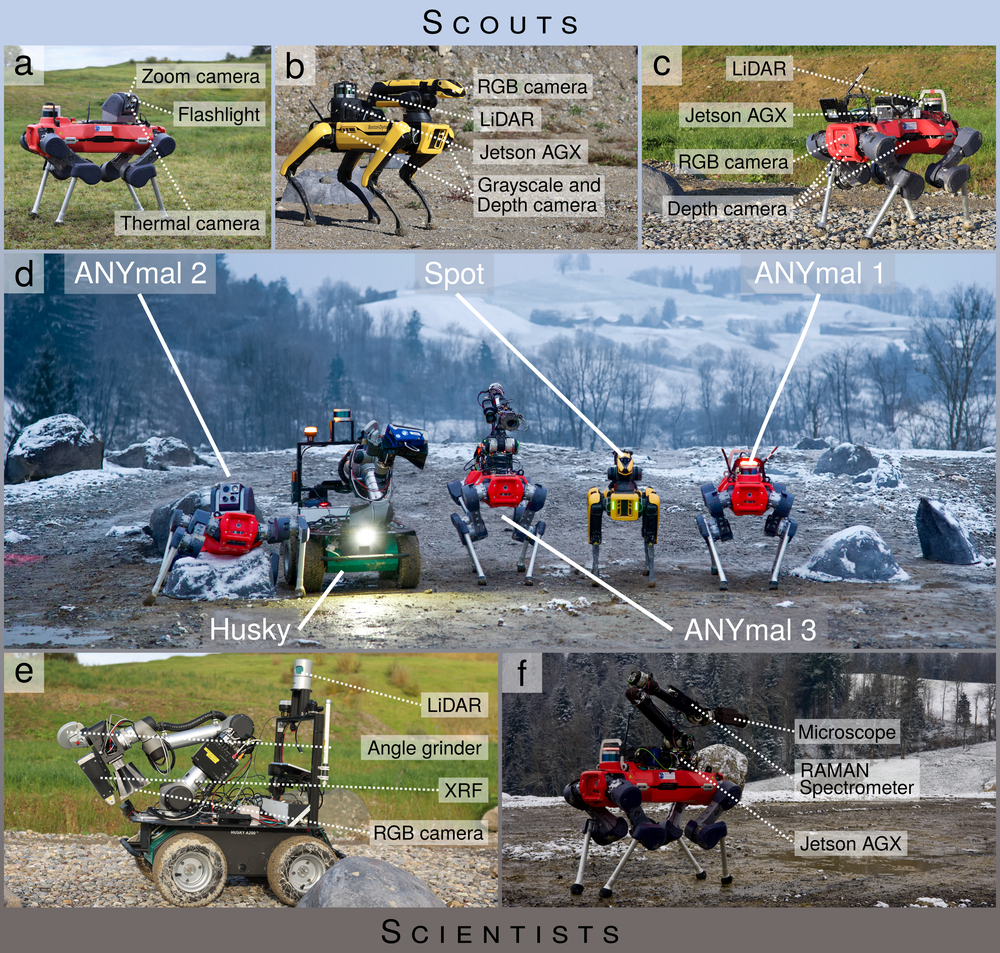}
    \caption{
    Overview of the heterogeneous robotic team used in this work.
    (a–c) Scout robots. 
    (d) Overview of the full deployed team. 
    (e–f) Scientist robots carrying manipulation and scientific payloads for close-range interaction and in-situ analysis.
    }
    \label{fig:robot_team}
\end{figure*}
\subsubsection{\anontext{ANYmal: Dodo}{ANYmal~1}}
\label{subsubsec: ANYMAL DODO}
\begin{figure*}[t]
    \centering
    \includegraphics[width=\textwidth]{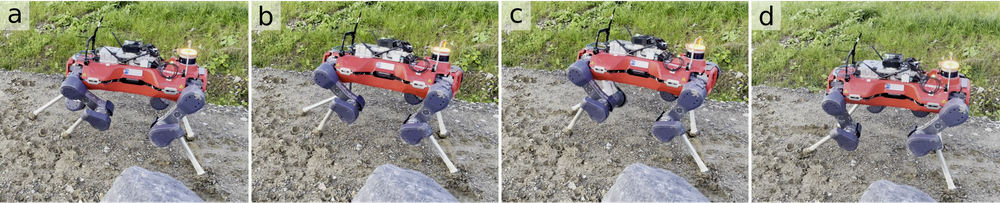}
    \caption{
        Trenching maneuver with the right front leg executed by \anontext{Dodo}{ANYmal~1} on loose terrain.
        (a-d) Consecutive frames illustrating the leg motions and body progression during the maneuver.
    }
    \label{fig:trenching}
\end{figure*}

The first scout robot is an ANYmal~D quadruped platform (see \cref{fig:robot_team}c), referred to as \anontext{Dodo}{ANYmal~1}.
Its sensor suite corresponds to the standard ANYmal~D configuration, comprising a 16-channel long-range Velodyne LiDAR, six short-range depth cameras, and two high-resolution wide-angle cameras.
In addition to the two onboard PCs, an NVIDIA Jetson AGX Orin is integrated to offload computationally demanding modules, including point cloud- and image-based rock detection.
Locomotion is enabled by a state-of-the-art perceptive policy based on reinforcement learning~\cite{miki_learning_2022}, which allows robust traversal of challenging terrain.
All robot-level modules, including body controllers, local navigation (\cref{subsubsec:local_planning}), and localization (\cref{subsec:shared_mapping}), are executed within ~\gls{ros1}. 
\paragraph{Pedipulation}
To enhance the robot's ability to identify promising candidates for scientists to investigate, we enable it to perform simple manipulation tasks, such as digging a small trench to expose subsurface material and flipping rocks to expose their underside, by utilizing its legs for manipulation.
One example for trenching is shown in~\Cref{fig:trenching}.
As the robot needs to perform these tasks on uneven, uncertain terrain, we rely on a reinforcement-learning whole-body controller to control the leg's end-effector position.
This tracking controller follows hand-designed trajectories for trenching and rock flipping~\cite{arm_pedipulate_2024}.

\paragraph{Rock Detection}
With the goal of enabling one operator to supervise three scouting robots, the volume of data to be analyzed in parallel is substantial.
Relying solely on the operator to identify \glspl{poi} is therefore infeasible, necessitating autonomous target detection. 
Given the available instruments, specifically the absence of long-range spectral cameras, the possible detection was limited to rocks rather than soil samples. 
Our approach combines a LiDAR-based clustering algorithm with a visual segmentation method based on a fine-tuned Detectron2 \cite{wu_detectron2_2019} model.
Since camera data alone is insufficient to determine the scientific relevance of a rock, we adopted a human-in-the-loop scheme: the robot proposes potential measurement targets, each consisting of one geometric obstacle and its associated segmented camera images, while the operator decides whether the target is of scientific interest. 
\subsubsection{\anontext{ANYmal~2: Dilly}{ANYmal~2}}
\label{subsubsec: ANYMAL DILLY}
The second scout robot, an ANYmal~D nicknamed \anontext{Dilly}{ANYmal~2} (\cref{fig:robot_team}a), is equipped with the same baseline sensor suite as \anontext{Dodo}{ANYmal~1} but additionally features the ANYbotics Inspection payload.
The payload includes a $ 20\times$ zoom camera, a thermal camera, an ultrasonic microphone, and a flashlight, all mounted on an actuated pan-tilt unit.
This payload enables the operator to manually inspect the robot's surroundings and perform in-depth inspections of potential~\glspl{poi}.
\anontext{Dilly}{ANYmal~2} uses the ANYbotics locomotion controller optimized for industrial use cases.
As with \anontext{Dodo}{ANYmal~1}, the locomotion controller and local navigation (\cref{subsubsec:local_planning}) are running in ~\gls{ros1}.
In addition, localization now also uses a \gls{ros2} module, as described in~\cref{subsec:shared_mapping}.
\subsubsection{Spot}
\label{subsubsec:SPOT}
The third scout robot is a Boston Dynamics Spot (\cref{fig:robot_team}b), a quadrupedal platform with an integrated manipulation arm.
It is equipped with five grayscale and five depth cameras, which the built-in software stack uses for obstacle avoidance.
Additionally, the arm is equipped with a high-resolution color camera, a depth camera, and a flashlight.
A custom-built payload extends the platform with a high-accuracy IMU and a Velodyne VLP-16 used for localization and navigation.
This payload also integrates an NVIDIA Jetson AGX Orin running a customized control stack.
Interaction with the Spot robot is facilitated via the \texttt{spot\_ros2} driver~\cite{noauthor_bdaiinstitutespot_ros2_2025}, which offers an interface between the proprietary Boston Dynamics API and \gls{ros2}.
All modules on Spot are running in ~\gls{ros2}.
\paragraph{Mobile Manipulator}
The robot is equipped with a 5-\gls{dof} arm and a rotatable two-finger gripper.
This arm can be controlled at the joint-level through the aforementioned~\gls{ros2} driver.
The arm is used under teleoperation to flip or inspect smaller rocks, similar to the pedipulation capability of \anontext{Dodo}{ANYmal~1}.
Teleoperation is performed using a 6D-Mouse, which enables precise positioning of the arm's end-effector.
Auxiliary teleoperation functions are implemented on top of the joint-level API, such as trajectory execution or autonomous stone picking based on pixel coordinates in the hand camera.
This is intended to allow an operator to perform complex manipulation with low cognitive overhead.
\subsection{Scientists}
\label{subsec:SCIENTISTS}
Scientist robots are designed to provide higher sensing and analytical capabilities than their scout teammates, typically carrying specialized payloads.
Their objective is to conduct detailed characterization and in-situ measurements once the scouts identify potential \glspl{poi}.
We employ two scientist platforms: an \textbf{ANYbotics ANYmal~D} quadruped robot equipped with a scientific payload for close-range measurements, and a wheeled \textbf{Clearpath Husky} carrying a modular sensor suite for long-duration surface analyses.
\subsubsection{\anontext{ANYmal~3: Donkey}{ANYmal~3}}
\label{subsubsec: anymal_donkey}
The first scientist robot is an ANYmal~D quadruped, designated \anontext{Donkey}{ANYmal~3}, shown in~\Cref{fig:robot_team}f.
Its sensor suite corresponds to the standard ANYmal~D configuration, and it is equipped with an NVIDIA Jetson AGX Orin for additional computing power.
The robot carries a 6-\gls{dof} manipulator arm, the DynaArm by Duatic \cite{duatic_ag_dynaarm_2025}, with a total weight of \SI{10}{kg}, including the \SI{2.3}{kg} scientific payload introduced in the following.
The arm has a reach of \SI{0.9}{m} and a maximum end-effector speed of \SI{10}{m/s}.
Control is split between two dedicated controllers~\cite{arm_scientific_2023}: locomotion is realized through a perceptive reinforcement learning policy~\cite{miki_learning_2022}, while the arm is operated through a model predictive control framework \cite{farshidian_ocs2_2025}.
\par
Equivalently to the other ANYmals, the robot-level modules (controllers, local navigation (\cref{subsubsec:local_planning}), localization (\cref{subsec:shared_mapping})) are implemented in~\gls{ros1}, while system-level functions (\gls{poi} planning (\cref{subsubsec:poi_planning}), global planning (\cref{subsubsec:global_planning}), shared mapping (\cref{subsec:shared_mapping})) are in~\gls{ros2}.
\paragraph{Scientific Equipment}
The manipulator is equipped with two scientific instruments: a Raman spectrometer and a microscope.
The Metrohm Mira XTR Raman spectrometer~\cite{noauthor_mira_nodate} features an autofocus function, enabling non-contact measurements from a distance of approximately \SI{0.7}{m}. It is mounted on the robot's forearm, allowing the end-effector to precisely maintain the required standoff distance.
The resulting Raman spectra can be matched against a mineral library to identify the mineralogy of a target sample.
The microscope is equipped with LEDs that cover the visible, near-infrared, and ultraviolet spectra, enabling microscopic imaging under various illumination conditions. To prevent interference from ambient light, the instrument requires controlled lighting conditions.
To this end, the microscope features a foam ring at the front that shields ambient light when the instrument is in contact with the target. The microscope is mounted directly on the end-effector to ensure accurate positioning.
\subsubsection{Husky} 
\label{subsubsec:husky}
The Clearpath Husky A200 (hereafter Husky, \cref{fig:robot_team}e) serves as a wheeled heavy-duty science platform.
Its wheel-based kinematics provide increased static stability and reduced energy consumption, particularly during stationary science operations.
Compared to the walking robots, Husky offers a substantially higher payload capacity, allowing the integration of a manipulator and additional sensors and tools.
The platform's static stability also enables force-intensive tasks, such as angle grinding.
This comes at the cost of slower movement speeds and reduced terrain adaptability.
\paragraph{Scientific Equipment}
Husky is equipped with a Universal Robots UR5 6-\gls{dof} arm with a payload capacity of \SI{5}{kg}.
This arm enables the use of scientific tools within a wide workspace, including operations on the ground and on nearby rock formations.
Built-in safety features restrict movements that could endanger the attached equipment or the robot during autonomous operation.
Attached to the arm's end-effector are three tools used for scientific operations.
First, an Olympus Vanta XRF~\cite{noauthor_vanta_nodate} is used for determining the metallic composition of a surface via~\gls{xrf}~\cite{grieken_handbook_2002}.
This method non-destructively identifies inorganic material compositions by analyzing the fluorescent X-rays emitted by materials after excitation with high-energy X-rays.
To reduce contamination of a surface by particles that have settled there and oxidation, it is beneficial to ablate the surface layer of sediment.
For this purpose, a high-power angle grinder is attached.
Lastly, a high-resolution RGB camera is used to photograph the ablated and measured surface close up for visual inspection.
To counteract adverse lighting conditions and ensure consistent lighting over multiple pictures, the camera is mounted in a cone with an internal light source.
\subsection{Autonomy}
\label{subsec: AUTONOMY}
In the following, we describe the~\mosaic scalable autonomy framework outlined in~\cref{subsec: AUTONOMY CONCEPT}, which builds upon the system presented in~\cite{schnell_efficient_2023}.
The approach is based on the principle that robots operate autonomously whenever possible, while allowing seamless transitions to lower levels of autonomy when human assistance is required.
Autonomy therefore adapts to the current mission context, the robot's state, and task complexity.
In practice, operator or scientist involvement remains essential for tasks that require expert judgment, such as selecting precise measurement locations, as well as for handling unexpected events or interruptions during mission execution.
Thus, we implement three control levels, which relate to the autonomy levels introduced in~\cref{subsec: AUTONOMY CONCEPT}:
\begin{itemize}
    \item \textbf{Driver-level control} is implemented via conditional access to direct teleoperation of each robot from the mission control station.
    \item \textbf{Task-level control} provides the operator with several tools:
        (i) directly commanding each robot to autonomously move to a specified target pose, (ii) triggering predefined behaviors, such as trenching pedipulation or specific arm trajectories, and (iii) providing information on suitable measurement positions and end-effector poses for selected \glspl{poi}, for which the robot awaited operator input, suspending its own mission-level autonomy.
    \item \textbf{Mission-level control} directly interacts with~\glspl{poi}, where different types of~\glspl{poi} correspond to different objectives executed by the robots as tasks.
\end{itemize}
In the following, we will describe how \glspl{poi} are selected and executed, which~\glspl{poi} exist, and how behavior trees are used to realize these missions.
\subsubsection{POI Selection}
\label{subsubssec:poi_selection}
\begin{figure*}[t]
    \centering
    {\sffamily \input{graphics/main_bt}}
    \caption{
        The main autonomy behavior tree for each robot in~\texttt{ros\_bt\_py}~\cite{heppner_distributed_2023}.
        On each robot, the subtrees handling the different POI types vary depending on the robot's capabilities.
        This tree runs for the entire duration of the mission.
    }
    \label{fig:main_bt}
\end{figure*}
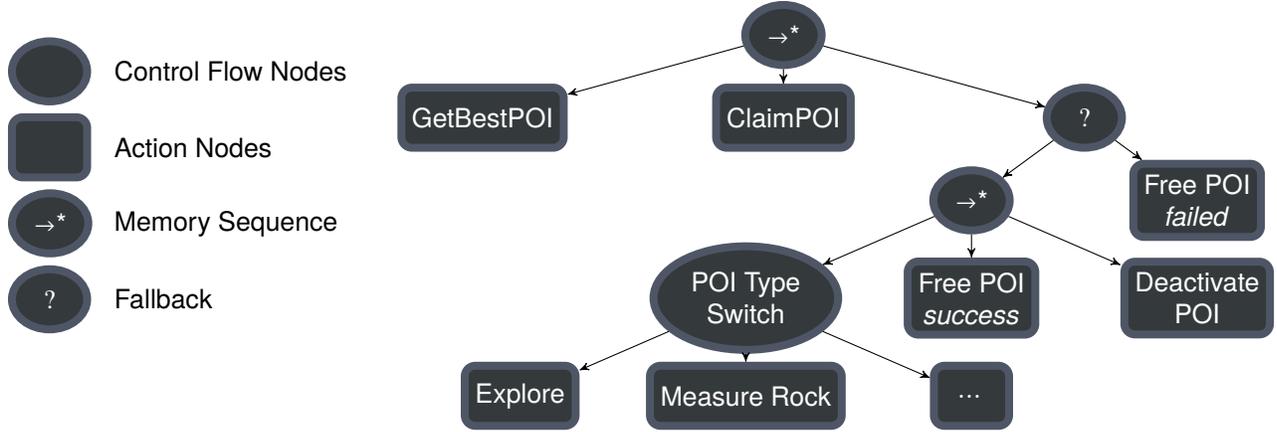
The main autonomy behavior tree (shown in~\cref{fig:main_bt}) realizes the main loop with the~\gls{poi} selection and execution behavior.
\Cref{alg:high_level_bt} showcases a pseudocode implementation of the functionality realized in the behavior tree.
\begin{algorithm}
\caption{Main behavior loop implemented in the high-level behavior tree for each robot.}
\label{alg:high_level_bt}
\begin{algorithmic}[1]
\Require $robot$ : \textsc{String}
\Require $PH$ : \textsc{Centralized POI Handler}
\Require $BS$ : \textsc{Base Station}
\While{true}
\State $teleoperation\_mode \gets$ \Call{${is\_teleoperated}_{BS}$}{robot}
\If{$\neg teleoperation\_mode$}
    \State $poi_{best} \gets $\Call{$get\_best\_poi_{robot}$}{~}
    \If{${poi_{best}} \not= none$}
        \State $result_{claim} \gets$ \Call{$claim\_poi_{PH}$}{$poi_{best}$, $robot$}
        \If{$\neg result_{claim}$}
            \State \Call{$record\_failure_{PH}$}{$poi_{best}$}
            \State \textbf{continue}
        \EndIf
        \State $result_{poi} \gets$ \Call{$execute\_poi_{robot}^{poi_{best}.type}$}{$poi_{best}$}
        \If{$result_{poi}$} 
            \State \Call{$record\_success_{PH}$}{$poi_{best}$, $robot$}
            \State \Call{$free\_poi_{PH}$}{$poi_{best}$}
            \State \Call{$deactivate\_poi_{PH}$}{$poi_{best}$}
        \Else
            \State \Call{$record\_failure_{PH}$}{$poi_{best}$, $robot$}
            \State \Call{$free\_poi_{PH}$}{$poi_{best}$}
        \EndIf
    \EndIf
\EndIf
\EndWhile
\end{algorithmic}
\end{algorithm}
Each robot selects an unclaimed~\gls{poi} via the planning process described in the following section and executes it using the appropriate subtree for the given POI type.
These subtrees vary in their exact implementation across robots in order to account for the heterogeneity of the system.
\subsubsection{POI Types} %
\label{subsubssec:poi_types}
Five~\gls{poi} types are used in our experiments:
\paragraph{Move}
A \texttt{MOVE} \gls{poi} wraps a simple navigation goal as a mission objective.
It is intended to guide the robot towards a specific position without performing any additional objectives.
This is particularly useful when the operator wants to position a robot at a certain location.
\texttt{MOVE} objectives represent the lowest utility for all robots in the mission.
\paragraph{Exploration}
An \texttt{EXPLORATION}~\gls{poi} extends a \texttt{MOVE} \gls{poi} by requiring the robot to rotate four times by $90^{{\circ}}$ at the target pose.
This ensures the entire surroundings are recorded on the map and provides more time for object detection.
For scout robots, this \gls{poi} type is prioritized, while for scientists it is ranked only slightly higher than  \texttt{MOVE} POIs.
\paragraph{Rock Candidate}
This~\gls{poi} type is generated by scouts when the \textit{rock detection} identifies a possible rock location.
A rock candidate is to be confirmed by a detailed visual inspection, preferably by a scout, which requires confirmation from the operator.
After which it is converted to a \texttt{Rock Measurement}~\gls{poi}.
\paragraph{Ground Measurement}
A \texttt{GROUND\_MEASUREMENT} \gls{poi} requires a scientist robot to conduct a measurement with its respective instrument on the ground surface beneath the specified pose.
For \anontext{Donkey}{ANYmal~3}, the type of measurement (Raman spectrometer, microscope, or combination of both) can be specified by the operator.
Husky always performs an XRF measurement of the ground.
\paragraph{Rock Measurement}
This~\gls{poi} type requires the robot to analyze the surface of a boulder or small rock.
Similarly to ground measurements, \anontext{Donkey}{ANYmal~3} can record any combination of its two sensors.
As outlined in~\cref{subsubsec:husky}, Husky grinds away the surface of the rock before recording the XRF measurement.
A~\texttt{ROCK\_CANDIDATE} can be converted to this \gls{poi} type if found suitable by the human scientists and operator.
\subsubsection{POI Planning}
\label{subsubsec:poi_planning}
To facilitate autonomous selection of~\glspl{poi}, we implement an iterative, multi-step greedy allocation algorithm that considers multiple metrics and other robots' preferred next steps.
Overall, the goal of the planning process is to efficiently allocate the available robot resources to suitable~\glspl{poi} with the intention that every~\gls{poi} will be completed during this mission.
The base metric used in the systems is called \textit{utility}, which represents the reward a robot receives for completing a specific objective.
Positive utility values indicate progress toward the mission, while negative values indicate infeasible assignments.
The planning is performed over a fixed number of steps, where each step selects a \gls{poi} based on the pose of the previous \gls{poi} (or the start position).
The planning process is structured into three components:
\paragraph{State Estimation}
The state estimation keeps a record of the predicted robot state along the planned~\gls{poi} execution.
After each planning step, it updates and stores the expected state for the next step.
The state primarily includes the robot's position but can also account for additional factors, such as battery consumption incurred by completing a \gls{poi}.
\paragraph{Utility Calculation:}
The utility calculation evaluates the utility ($u_r(s_r, p)$) of each unclaimed~\gls{poi} (p) from the current state estimate ($s_r$) of each robot (r).
\begin{equation}
    u_r(s_r, p) = \sum_{f_r \in \text{Utility Features}} w_{f_r} * f_r(s_r, p)
\end{equation}
It is implemented as a plugin-based system that allows for an arbitrary number of \enquote{\textit{utility-features}} ($f_r(s_r, p)$) to be loaded and considered for a \gls{poi}'s utility.
Some features reward (\uparrow, $w_{f_r} * f_r(s_r, p) >= 0$) the robot for the~\gls{poi}, whereas others feature model penalties (\downarrow, $w_{f_r} * f_r(s_r, p) <= 0$).
Common features include:
\begin{itemize}
    \item \textbf{Type of POI}  (\uparrow): Gives a fixed positive reward or zero depending on the robot's ability to complete this POI type.
    This reward is pre-configured by the operators per POI type and robot based on prior assessments of capabilities.
    \item \textbf{Euclidean Distance} (\downarrow): Euclidean distance of the robot's current (or planned) position to the POI. 
    \item \textbf{Navigation Distance} (\downarrow): Length of the path the navigation estimates to the position of the POI.
    \item \textbf{Battery Consumption} (\downarrow): This feature is dependent on one of the distance metrics and accounts for the consumption of battery along the given path, together with the currently available charge.
\end{itemize}
These (robot-intrinsic) features were selected to prioritize the timely completion of~\glspl{poi} by suitable robots, not to determine the~\gls{poi}'s mission value.
While features for estimating the mission value of a~\gls{poi} at runtime could be implemented through the plugin architecture, this was out of scope for this work due to the inherent complexity of calculating such a measure.
All feature utilities are weighted by robot- and mission-specific weights ($w_{f_r} \in [-1.0, 1.0]$), which the operator can configure to adapt the planner's behavior to different deployment scenarios.
The total utility for a poi and robot ($u_r(s_r, t)$) is the sum of the weight features.
\par
The \gls{poi} planning requires two types of utility calculations:
First, an estimation of the \textbf{maximum possible utility} ($u^{max}_r(s_r, p)$) is computed using the Euclidean distance only, ignoring navigation and battery constraints.
Second, a fine-grained calculation ($u^{det}_r(s_r, p)$) refines the estimate by incorporating navigation distance and battery consumption.
By splitting the utility calculation into these two steps, the costly navigation and battery calculations are only performed for the most promising \glspl{poi}, while distant candidates are pruned early.
\par
This design is intended to provide a flexible system for defining relevant metrics to assess a~\glspl{poi} contribution to the overall mission progress, while still permitting interpretation of the estimates and intuitive adjustments by an operator.
By estimating mission progress solely from the robot state and~\gls{poi} attributes, the estimation process is purely encapsulated within each robot and can be computed efficiently.
\paragraph{POI Planning}
The two introduced components are coordinated by the POI planning component, which selects the most suitable POI for a robot by evaluating and comparing candidate POIs based on their expected utility and coordination with other robots.
For each planning step up to a predefined planning depth, the module recursively selects the next POI, resulting in an ordered list of POIs that constitutes a plan.
The robot will then execute the first POI of this plan.
\par
The POI planning starts by retrieving the current list of \glspl{poi} ($\mathcal{P}$) from mission control for goal selection.
Initially, the maximum possible utility for each~\gls{poi} is calculated based on an operator-defined, less computationally intensive set of utility-features that provide an upper estimate.
\begin{equation}
    \begin{split}
        \mathcal{P}' = \operatorname{sort}_{u^{max}_r(s^0_r)}\!\Bigl(\bigl\{ p \in \mathcal{P} \;\big|\; 
            &u^{max}_r(s^0_r, p) \geq 0 \\
            &\wedge\; u^{max}_r(s^0_r, p) \geq \max_{r' \neq r} u_{r'}(p)
        \bigr\}\Bigr)
    \end{split}
\end{equation}
\glspl{poi} with a negative estimated maximum utility, as well as those for which another robot has already registered a higher utility, are filtered out.
For the remaining \glspl{poi} ($\mathcal{P}'$), a detailed utility calculation ($u^{det}_{r}(s^0_r, p^0)$), utilizing more exact but computationally costly utility features, is performed.
The resulting values are weighted according to the current planning depth ($D$) using:
\begin{equation}
    \begin{aligned}
    \mathcal{P}'' &= \langle p^*_0, p^*_1, \ldots, p^*_D \rangle \\
    \text{with:}& \\
    p^*_0 &= \mathcal{P}'[0], \\
    s^{d+1}_r &= S(s^d_r, p^*_d),\\
    p^*_{d} &= \arg\max_{p \in \mathcal{P}' \setminus \{p^*_0, \ldots, p^*_{d-1}\}} u^{det}_r(s^d_r, p) \cdot \delta^{d}, \quad d \geq 1
    \end{aligned}   
\end{equation}
with $\delta$ being the depth uncertainty factor.
This value $\delta \in (0.0, 1.0]$ is selected to represent the inherent uncertainty of planning into the future.
Lower values of $\delta$ decrease the overall utility gained from~\gls{poi} later in the plan.
This is to account for unknown factors (e.g., unknown terrain features) and incentivizes robots to complete~\glspl{poi} more promptly, without being blocked by robots who have them late in their plans.
The final list of utilities is sorted in decreasing order.
The first \gls{poi} ($p^*_0$) from that list, which has a positive utility and no higher utility recorded from another robot, is selected as the best \gls{poi} for this planning step.
After each planning step, the state estimation ($S(\dots)$) is invoked to calculate the next state of the robot.
The weighted utility $u^{det}_r(s^d_r, p) \cdot \delta^{d}$ is recorded in the \gls{poi} handler for this \gls{poi} ($"p^*_{d}$) and robot, and the process is repeated until the desired planning depth is reached.
By planning not just to the next~\gls{poi}, but also a few steps ahead, we can, for example, avoid choosing plans that yield high utility now and significantly lower utilities in the future.
The resulting plan ($\mathcal{P}''$) is then shared with the other robots to inform their planning.
The function ultimately returns the \gls{poi} selected in the first planning step as the robot's best \gls{poi} ($p^*_0$).
\par
This greedy algorithm enables rapid~\gls{poi} plan generation for each robot, accounting for multiple metrics and teammate preferences while maintaining low communication overhead and supporting rapid replanning in response to operator interventions or unexpected events.
The resulting plans constitute sufficient approximations given the operational constraints of the target mission class; for missions requiring provably optimal resource utilization over extended durations, more sophisticated planners would be better suited.
\subsection{Mapping and Localization} %
\label{subsec:shared_mapping}
One key factor in multi-robot team setups is the ability to create shared maps and establish co-localization, ensuring that all robots have a consistent view of the shared workspace.
This constant reference frame also simplifies the operator's interaction and decision-making with the whole team.
\paragraph{Shared Mapping}
We utilize the \texttt{vdb\_mapping}~\cite{grosse_besselmann_vdb-mapping_2021} framework for volumetric 3D mapping on each robot.
Each individual map is generated from the long-range~\gls{lidar} and the short-range depth cameras of the respective robot.
This map is used for autonomous navigation on the robot, as explained in further detail in the following section.
\par
To build a unified global map, each robot periodically transmits a local map section (a  \qty{20}{\meter}~x~\qty{20}{\meter}~x~\qty{20}{\meter} cube) to mission control, where the sections are merged according to the robot's current position in the shared \textit{earth} coordinate system.
This shared frame is created at startup using Iterative Closest Point (ICP) alignment of map sections acquired while the robots are in close proximity to each other.
After approximate initialization by the operator, the ICP algorithm refines the alignment.
The resulting coordinate transformations are subsequently used to align newly incoming segments during the exploration process.
\paragraph{Co-localization}
\label{par:co-localization}
After the initial alignment of the robot maps, no continuous co-localization is performed.
Instead, the system relies on the initial ICP estimate.
Each robot subsequently maintains its own localization using platform-specific methods. 
For \anontext{Dodo}{ANYmal~1} and \anontext{Donkey}{ANYmal~3}, we deployed Open3D SLAM~\cite{noauthor_leggedroboticsopen3d_slam_2025}.
\anontext{Dilly}{ANYmal~2}, Husky and Spot rely on the Google Cartographer SLAM~\cite{hess_real-time_2016} system, utilizing~\gls{lidar}, odometry, and~\gls{imu} data.
\subsection{Navigation} %
All robots employ the NavPi framework for 3D navigation~\cite{grosse_besselmann_3d_2024}.
The navigation pipeline is structured into two stages: a global planning step followed by a local planning step.
While the global planner is identical across all platforms, the local planner is tailored for each robot to accommodate its specific locomotion requirements.
\subsubsection{Global Planning} %
\label{subsubsec:global_planning}
The goal of the global planner is to find a coarse path from the start position to the goal pose.
This plan is created on a static snapshot of the robot's \texttt{vdb\_mapping} map at the start of the planning process.
NavPi's plugin-based architecture enables the use of various volumetric 3D planners to generate three-dimensional paths.
\par
The first option is an A* search, which generates optimal paths at the expense of increased computation time.
As the search space becomes more complex with path length, A* becomes infeasible for long-range planning.
However, it is useful for short paths in complex terrain, where an optimal path may improve stability and safety.
\par
The second option uses sampling-based planners from the Open Motion Planning Library (OMPL), which NavPi integrates as plugins.
Due to their probabilistic nature, they often sacrifice optimality for significantly reduced planning times.
Since the generated paths remain collision-free, these planners are well-suited for planning longer paths through easier terrain.
\par
Having the option to switch between planners on demand increases the system's flexibility and supports situation-aware navigation.
By default, the RRT Connect OMPL sampling-based planner is used, with the other OMPL-based planners and A* serving as situational alternatives.
Once a global path is found by any planner, it is discretized into sub-goals, which are subsequently passed to the local planning step.
\subsubsection{Local Planning}
\label{subsubsec:local_planning}
As the robots utilize different modes and methods of locomotion, the local planner components are adapted to allow the optimal utilization of each platform's capabilities.
\paragraph{ANYmal~D}
The three ANYmals use the Field Local Planner~\cite{mattamala_efficient_2022, noauthor_ori-drsfield_local_planner_2025}, which optimizes local paths using the ANYmals 2.5D local elevation map.
This map is built from~\gls{lidar} and depth camera data. Leveraging the onboard GPUs, for \anontext{Dodo}{ANYmal~1} and \anontext{Donkey}{ANYmal~3}, elevation maps are produced with the GPU-accelerated Elevation Mapping Cupy framework~\cite{miki_elevation_2022, erni_mem_2023, noauthor_leggedroboticselevation_mapping_cupy_2025}, while \anontext{Dilly}{ANYmal~2} relies on ANYbotics’ Near Field Mapping system.
This is unrelated to the global 3D map used for the global navigation.
\par
From the elevation map, the local planner derives a Signed Distance Field (SDF) for obstacle avoidance and a Geodesic Distance Field (GDF) for goal-directed motion. 
The fields are fused using Riemannian Motion Policies (RMP), enabling the robot to walk to the lookahead goal point while avoiding obstacles. 
The motion commands are generated at \SI{10}{Hz}, ensuring fast and responsive local navigation and allowing the robot to replan around dynamic obstacles, including other team members.
\paragraph{Spot}
On Spot, the proprietary local planner is used and integrates seamlessly with the rest of the navigation stack.
For this, a close target pose is sampled from the global path and converted into a valid input for the Spot API.
Subsequently, the local planner guides the robot toward this intermediate target pose while avoiding obstacles.
As the system is proprietary to Boston Dynamics, we are not able to provide additional details.
\paragraph{Husky}
Husky utilizes the global planner NavPi for both local and global planning~\cite{grosse_besselmann_navpi_2024}.
As with the other robots, a local lookahead goal is selected and given to the planner. 
The planner creates multiple directional fields, which provide the most suitable possible path towards the intermediate goal while avoiding obstacles. 
\subsection{Operator Workflows}
The human operators, both robot operators and science experts, are a central component of~\mosaic.
They observe and occasionally intervene during autonomous operations.
As outlined in~\cref{sec:RELATED_WORKS}, our approach aims to minimize the operator workload to ensure scalability with respect to team size.
Accordingly, we optimize the workflows for both the main operator, who is responsible for the mission control station (\Cref{fig:base_station_ui:base_station}), and the scientific operators, who interact with the science operations station (\Cref{fig:base_station_ui:science_station}).
The main operator oversees team management and teleoperation of the robotic platforms~\cref{sec:team_management}.
The scientific operators analyze information gathered by scouts and scientist robots and perform the detailed selection of measurement targets.
\subsubsection{Team Management Workflow}
\label{sec:team_management}
Team management includes the active management of~\glspl{poi}, troubleshooting, fault clearing, and mission progression tracking.
To provide the operator with an up-to-date view of the robots' surroundings, we used a 6-screen setup (\Cref{fig:base_station_ui:base_station}) with a 3D voxel-based global map and each robot's current configuration as the primary information source.
As a secondary interface, we streamed images of each robot's front cameras to the operator.
Lastly, status information, such as battery or motor states, is available in the individual robot screens.
We realize \Gls{poi} management via interactive markers in the 3D Environment visualization as shown in the \textit{Map \& POI Planning} view.
A \gls{poi} can be created, moved, altered, assigned, and deleted by the operator using these markers.
Completed~\glspl{poi} are not automatically removed, but set transparent, to track past mission progress and allow reactivating them should follow-up measurements become necessary.
To alter the robot's autonomous behavior, the operator has live access to each behavior tree running on the individual robots.
At the robot level of the autonomy concept, the operator can utilize various helper functions, such as executing specific arm functions or trajectories, as well as altering autonomous navigation goals.
For driver-level access, we used a 6D-Mouse, allowing the fine-grained teleoperation of each robot's base, as well as end-effectors.
\subsubsection{Scientific Measurement Workflow}
During the mission, data from scouts and scientist robots—including camera images, scientific measurements, and the shared 3D map—are synchronized to both the mission control and science operations stations.
The science operations station enables geology experts to review incoming data, assess measurement quality, and identify missing or follow-up observations.
This feedback is communicated to the operator and informs the prioritization of subsequent objectives.

Because identifying suitable measurement locations for the scientist robots’ instruments is challenging to automate reliably, we employ a human-in-the-loop workflow.
When a robot reaches a measurement pose, it transmits an image of the target area to the science operations station, where the researcher selects the precise measurement point using a 2D interface.
The operator then refines the corresponding 3D end-effector pose to ensure it satisfies both instrument and operational constraints.
This division of responsibilities allows researchers to focus on scientific decision-making, while the operator manages execution- or robot-specific constraints, like joint limits or reachability.
\subsection{Communication System}
A core challenge when operating a multi-robot team is the communication between agents.
As mentioned in \cref{sec: SYSTEM ARCHITECTURE}, we chose~\gls{ros2} Humble with FastDDS as the application framework and middleware for~\mosaic.
This choice offers several key benefits: avoiding a central communication master, configurable~\gls{qos} for different topics, and a clear separation between shared and robot-specific communication.
FastDDS, the chosen~\gls{dds} implementation, relies on UDP/IP messaging between hosts and employs shared memory for data transfers on the same device.
At the physical layer, we use Wi-Fi 5 for communication between the robots and the mission control router, due to its high bandwidth and straightforward setup.
Other wireless physical layers are possible, as the stack does not depend on the actual physical layer used.
All robots operate in a shared subnet for team-level communication.
Additionally, each robot has an internal network for communication between its various internal computers.
To mirror this on the~\gls{ros2} side, we define a shared \texttt{ROS\_DOMAIN\_ID} for team-level communication, alongside a dedicated domain for each robot.
Communication between different \texttt{ROS\_DOMAIN\_ID}s is handled by a domain bridge, which relays a configured set of topics between different~\gls{ros2} domains, while altering some~\gls{qos} settings to better account for the shared nature of the main domain.
\paragraph{Quality of Service}
One key feature of~\gls{ros2} is its ability to make fine-grained adjustments to the transmission behavior for each topic via~\gls{qos} settings.
To ensure minimal usage of the shared bandwidth between the robots, we adjusted the~\gls{qos} by dividing topics into three broad categories: control, sensor, and transforms (TF).
Control topics (e.g., twist commands, BT commands) are generally transmitted as \textit{reliable} to guarantee delivery, while sensor topics are designated \textit{best effort}, tolerating occasional packet loss.
An important observation was that for unreliable links such as Wi-Fi,~\gls{ros2} exhibits a more stable transmission behavior if all messages are sent with~\textit{best effort}, as this avoids retransmission backups that negatively affect latency.
Maintaining a valid transform tree is essential for many~\gls{ros1} and~\gls{ros2} autonomy modules.
Static (unchanging) transforms are transmitted reliably with \textit{transient local} durability, allowing late-joining subscribers to receive them.
Dynamic transforms, published regularly, are submitted with \textit{best effort}.
To avoid bandwidth saturation during high-rate publishing transforms, it was necessary to throttle the message rate to below \SI{100}{\Hz}.
To prevent transform loss due to publishers operating at different frequencies, this was achieved using a \textit{smart transform throttle}, which aggregates transforms, removes duplicates, and republishes them at a fixed rate across the domain bridge. 
\paragraph{\gls{ros1} $\leftrightarrow$ \gls{ros2} Bridge}
Because ANYmal's internal drivers rely on \gls{ros1}, a communication bridge between \gls{ros1} and \gls{ros2} is required.
This bridge is configured for bidirectional mapping between certain topics, services, and actions.
To reduce network utilization and consequently latency~\cite{maruyama_exploring_2016}, both the number of topics and the publishing frequency are limited.
Similar to the domain bridge setup, we separated control, sensing, and transforms into separate bridge instances for increased stability.
\paragraph{Syncing of Mission Data}
Collected scientific data must be transferred to the science operations and mission control stations during mission runtime.
This is not handled via~\gls{ros2}-based communication, to avoid sending large data chunks, possibly congesting control traffic.
Instead, we use \texttt{rsync} to synchronize files from each robot to the mission control station and, subsequently, to the science operations station.
Recorded data are compressed before transfer.
Failed sends are automatically retried, along with newly recorded data, to minimize parallel connections.

%% file: graphics/main_bt.tex
\begin{tikzpicture}[->,>=stealth',
            level 1/.style={sibling distance=4cm, level distance=1.1cm},
            level 2/.style={sibling distance=3cm, level distance=1.1cm},
            level 3/.style={sibling distance=3cm, level distance=1.3cm}]
            \node [btcondition] (top_level_seq) {$\rightarrow$*}
            child { node [btaction] (get_best_poi_1) {GetBestPOI}}
            child { node [btaction] (claim_poi) {ClaimPOI}}
            child { node [btcondition] (fallback) {$?$} 
                child { node [btcondition] (poi_seq) {$\rightarrow$*}
                    child { node [btcondition] (name_switch) {POI Type \\ Switch}
                        child {node [btaction] (expore_poi) {Explore}}
                        child {node [btaction] (measure_poi) {Measure Rock}}
                        child {node [btaction] (dots_poi) {$\dots$}}
                    }
                    child { node [btaction] (free_poi_success) {Free POI \\ \textit{success}}}
                    child { node [btaction] (deactivate_poi) {Deactivate \\ POI}}
                }
                child { node [btaction] (free_poi_failed) {Free POI\\ \textit{failed}}}
            };
            \matrix[matrix of nodes,
                anchor=east,
                column sep=5pt,
                nodes={anchor=west},
                row sep=3pt
            ] (legend) at ([xshift=-5cm, yshift=-2cm]top_level_seq.west) {
                \node[btcondition] {~~~};~: & Control Flow Nodes  \\
                \node[btaction] {~~~};~: & Action Nodes  \\
                \node[btcondition] {$\rightarrow$*};~:& Memory Sequence \\
                \node[btcondition] {$?$};~: & Fallback \\
            };
        \end{tikzpicture}

%% file: sections/05_experiments.tex
\begin{figure*}[t]
    \centering
    \includegraphics[width=\textwidth]{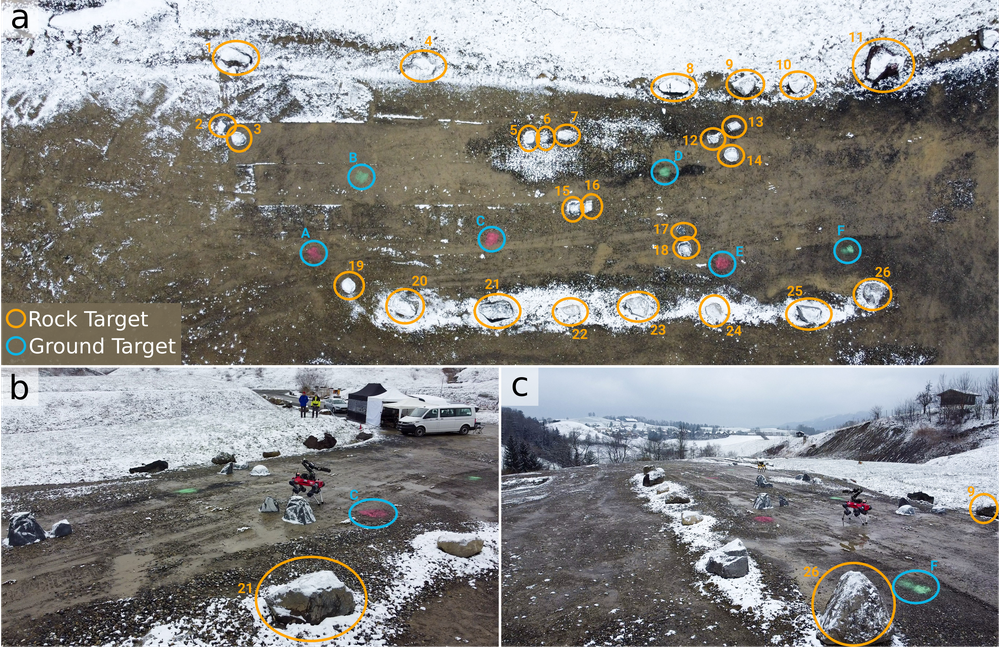}
    \caption{
        Overview of the experimental test site and sampling targets.
        a) Target map derived from drone imagery, indicating measurement targets across the site. Rock targets are indicated by \textcolor{c_rock_markers}{orange circles}, while ground targets (colored sand patches) are marked in \textcolor{c_sand_markers}{blue}.
        b-c) Representative ground-level views of the test site with targets marked for correspondence with the aerial map.
    }
    \label{fig:neuheim_target_map}
\end{figure*}
\section{FIELD EXPERIMENTS}
\label{sec:EXPERIMENTS}
Field deployments are essential for assessing multi‑robot systems, as many aspects of system behavior, such as robot coordination, robustness to unexpected events, and the influence of environmental conditions, only become apparent outside of simulation and controlled experiments.
In order to test our~\mosaic framework, we conducted a field test inspired by a lunar prospecting scenario.
In the following, we describe the mission scenario and its rationale (\cref{subsec:mission design}), the physical test environment and target layout (\cref{subsec:testing area}), and the operational procedure used during the experiments (\cref{subsec:experimental design}).
\subsection{Mission Design and Rationale}
\label{subsec:mission design}
Lunar resource prospecting is a representative example of multi-robot autonomy, as robots must operate in unknown terrain, generate new objectives during the mission, and balance broad exploration with targeted measurements. 
Delayed or interrupted communication further requires a high level of onboard autonomy in both task allocation and navigation.
\par
For these reasons, the scenario provides a suitable context for evaluating our scalable autonomy framework. Based on this motivation, the field mission focused on detecting resource proxies such as boulders and regolith-like patches, with the operator supervising from a mission control station without direct line of sight and managing points of interest through a unified interface.
\subsection{Test Environment and Target Layout}
\label{subsec:testing area}
The experiments were conducted at a quarry near Neuheim, Switzerland, which resembles a lunar analog scenario due to its degraded environment and loose soil.
The terrain comprises a predominantly flat central area bordered by steeper flanks, as seen in~\cref{fig:neuheim_target_map}.
The active test area was approximately $60~\mathrm{m}\times 40~\mathrm{m}$.
The tests were conducted in February, resulting in unfavorable weather conditions.
Specifically, snow and wet soil increase slippage and sinkage, making locomotion more challenging than in dry conditions.
\par
To emulate resource-prospecting targets, we deployed two types of proxies. 
Natural boulders were positioned along the upper and lower edges of the site, while additional plastic boulder surrogates were placed throughout the central region.
To simulate mineralogical diversity, we introduced colored sand patches in red, green, and blue tones. 
Although real \gls{ree} deposits are typically identified using multispectral instruments, we used RGB detection for these experiments due to sensor availability. 
A complete overview of the target distribution and aerial views of the site are shown in \cref{fig:neuheim_target_map}.
\subsection{Operational Procedure}
\label{subsec:experimental design}
Each mission began at a simulated lander site, with all robots deployed from a common starting area. 
The operator’s responsibilities were defined in accordance with the goals of the scalable autonomy framework. 
Rather than relying on autonomous frontier exploration, the operator guided global mission progress by placing and refining \glspl{poi}. 
This approach leverages human intuition to incorporate global context and identify distant terrain features that are difficult to capture with onboard sensing and automated perception pipelines.
In addition, while the system relied on the scout \anontext{Dodo}{ANYmal~1} to detect potential scientific targets, the operator's task was to validate and adjust these suggestions and add further \glspl{poi} based on live camera feeds from all robots.
Beyond exploration planning, the operator was also responsible for mission-level risk management.
This included directly teleoperating a robot in the event of autonomy failure and avoiding the redundant deployment of multiple robots into the same high-risk zone, thereby distributing risk across the team.
\par
The overall mission objective was to generate a consistent three-dimensional map of the environment enriched with detected \glspl{poi} and associated measurements. 
Relevant data included multi-view camera observations and platform-specific sensor readings. 
All robot telemetry, \gls{poi} updates, and operator interactions were logged for post-mission analysis of multi-robot coordination, autonomy behavior, and supervisory workload.

%% file: sections/06_evaluation.tex
\section{RESULTS}
\label{sec:RESULTS}
\begin{table*}
  \caption{\Acrfullpl{kpi} for multi-robot lunar missions, reported from the final field test with all robots deployed.}
  \label{tab:kpis}
  \centering
  \footnotesize
  \setlength{\tabcolsep}{3pt}
  \renewcommand{\arraystretch}{1.1}
  \begin{tabular}{@{}p{2.2cm} p{6cm} l p{2cm}@{}}
    \toprule
    \textbf{Category} & \textbf{KPI} & \textbf{Unit} & \textbf{Value} \\ \midrule[1pt]
    
    \multirow{5}{=}{Efficiency}
        & Total Explored Area & \si{\meter\squared} & 758 \\ \cmidrule(lr){2-4}
        & Mapping Efficiency & \si{\meter\squared\per\meter} & 2.19 \\ \cmidrule(lr){2-4}
        & Mapping Rate & \si{\meter\squared\per\second} & 0.313 \\ \cmidrule(lr){2-4}
        & Task Success Ratio & \% & 82.3 \\ \cmidrule(lr){2-4}
        & Quantitative Operator Workload & \% & 78.2 \\ \midrule[1pt]

    \multirow
    {2}{=}{Robustness}
        & Robot Downtime & \% & 52 \\ \cmidrule(lr){2-4}
        & Robot Downtime (excluding Husky due to incorrect state labeling) & \% & 37.8 \\ \cmidrule(lr){2-4}
        & Autonomy Ratio & \% & 86\\ \cmidrule(lr){2-4}
        & Time in Unscheduled Manual Robot Operations & \% & 8.45 \\ \cmidrule(lr){2-4}
        & Retry Ratio & \% & 19.4 \\ \midrule[1pt]
        
    \multirow{1}{=}{Precision}
        & Map Error & \si{\meter} & 0.145 \pm 0.252 \\ \cmidrule(lr){2-4}
        & Ratio of Identified Resources & \% & 56.2 \\ 
    \bottomrule
  \end{tabular}
  \vspace{-1.3em}
\end{table*}

\begin{figure*}[t]
    \centering
    \subfloat[Aerial image overlaid with robot trajectories and measurement locations.\label{fig:mission_paths:paths}]{
        \includegraphics[
            width=0.85\textwidth,
            trim=0 240mm 0 0,    %
            clip
        ]{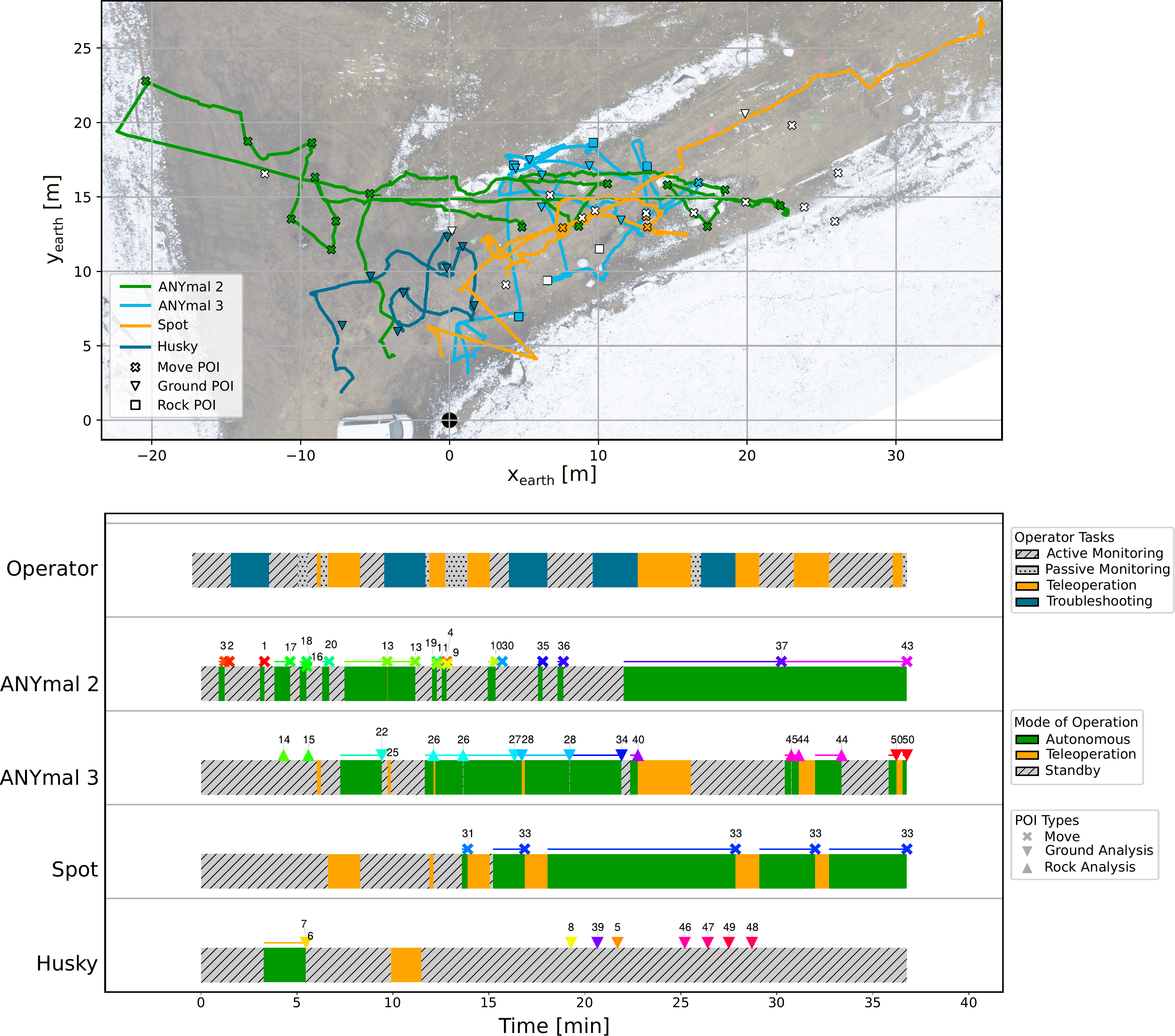}
        }\\
    \subfloat[Operator activity timeline and per-robot task execution with autonomy modes.\label{fig:mission_paths:timeline}]{
        \includegraphics[
            width=0.85\textwidth,
            trim=0 0 0 220mm,    %
            clip
        ]{images/mission_run.pdf}
        }
    \caption{
        Mission overview, including trajectory overview (a) and activity timeline (b). 
    }
    \label{fig:mission1_paths}
\end{figure*}
Across several field campaigns from October 2024 to February 2025,~\mosaic was iteratively tested and refined. 
The results presented here refer to the final campaign, which involved the deployment of all functioning robots.
\par
To evaluate the mission outcome, we summarize the mission's overall qualitative performance alongside a quantitative analysis based on a subset of the~\gls{kpi} framework proposed in~\cite{richter_practical_2026}.
Specifically, we selected~\glspl{kpi} focusing on efficiency, robustness, and scientific precision under realistic deployment conditions.
This~\gls{kpi} selection reflects the contributions of both robot roles: exploration-related metrics capture the performance of scout platforms, while measurement-related metrics characterize the effectiveness of scientist platforms.
The results of the~\gls{kpi} analysis are reported in~\Cref{tab:kpis} and are discussed in the following sections.

\subsection{Operational Performance and Autonomy Analysis}
This section analyzes mission execution from an operational perspective, focusing on autonomy levels, operator workload, and system robustness.
The mission successfully accomplished its primary objectives, demonstrating the feasibility of coordinated multi-robot exploration, while several unforeseen events provided insight into system robustness and operator supervision requirements.
\par
During the first two days of the test campaign, the scout robot \anontext{Dodo}{ANYmal~1} suffered water damage and had to be taken out of operation, removing its autonomous rock-detection capability.
The mission could be continued without major disruption.
However, this significantly increased the \textit{Quantitative Operator Workload} (78.2\%), especially the time spent on active monitoring (27.3\%), as \glspl{poi} had to be created manually.
\Cref{fig:mission_paths:timeline} provides a detailed breakdown of operator activities over the course of the mission.
\par
During the first part of the mission, Spot additionally encountered a software and networking issue that prevented it from running its autonomous behavior tree. 
During troubleshooting, it was used as a mobile camera platform, supporting operator oversight and helping to guide \anontext{Donkey}{ANYmal~3} during the first measurement tasks.
Troubleshooting various robots, especially Spot, accounted for around 19.5\% of the \textit{Quantitative Operator Workload}.
Additional time was required for the teleoperation (31.4\%), mainly for Spot (17.0\%).
Given the aforementioned manual \gls{poi} generation and the technical issues, the system still maintained a relatively high overall \textit{Autonomy Ratio} of 86\%.
This metric is defined as the inverse of the Robot Attention Demand (RAD), computed as $RAD=\tfrac{IE}{IE+NT}$, where $IE$ denotes the interaction effort, measured by the time spent on operator teleoperation and troubleshooting, and $NT$ represents the neglect tolerance, defined as periods during which the robots autonomously executed POIs or performed POI planning.
\par
The \textit{Time in Unscheduled Manual Robot Operations}, defined as the fraction of mission time during which robots required teleoperation, remained low at 8.45\% overall but showed considerable variation across platforms.
Husky required manual intervention for only 4.3\% of the mission time, while \anontext{Dilly}{ANYmal~2} operated almost fully autonomously, with teleoperation accounting for just 0.1\%.
In contrast, \anontext{Donkey}{ANYmal~3} required manual control for 12.4\% of the mission, primarily to support fine positioning prior to automated scientific measurements.
Spot showed the highest reliance on teleoperation at 17.0\%, reflecting the previously discussed issues in behavior tree execution.

To further assess robot utilization, we analyze \textit{Robot Downtime}, defined as periods during which robots were not actively contributing to mission progress by executing POIs.
This includes initial idle time before the behavior tree was started, for instance, due to startup delays or malfunctions, as well as time spent in planning states between tasks. Teleoperated execution of POIs is explicitly considered active time and therefore excluded from downtime.

Across the mission, the cumulative robot \textit{Robot Downtime} amounted to 52\%.
This relatively high value is primarily driven by extended standby periods of Husky, as illustrated in \Cref{fig:mission_paths:timeline}.
Although Husky was actively executing POIs, its behavior tree failed to correctly register task execution, resulting in an apparent planning time of 80.8\% and an idle time of 8.9\%.
Since the robot continued to perform meaningful work during these intervals, we attribute this downtime largely to incorrect state labeling, likely caused by a missing claim confirmation in the behavior tree, rather than actual inactivity.
Excluding Husky from the analysis reduces the planning-related downtime of the remaining robots to 24.1\% and the total \textit{Robot Downtime} to 37.8\%.

A second major contributor to the overall downtime was Spot, which remained inactive for 31.8\% of the mission time due to technical issues at the start of the deployment. 
In comparison, \anontext{Dilly}{ANYmal~2} and \anontext{Donkey}{ANYmal~3} showed substantially lower idle times of 2.5\% and 11.7\%, respectively, with planning-related downtime of 37.6\% and 34.3\%.
These results indicate that the observed downtime is largely driven by platform-specific issues and state-estimation artifacts, rather than a fundamental lack of autonomy during task execution.
Across the mission, the total \textit{Downtime} comprised 13.7\% idle time and 38.3\% planning time, highlighting planning efficiency as a key area for future improvement.

\begin{figure*}[t]
    \centering
    \includegraphics[width=.9\linewidth]{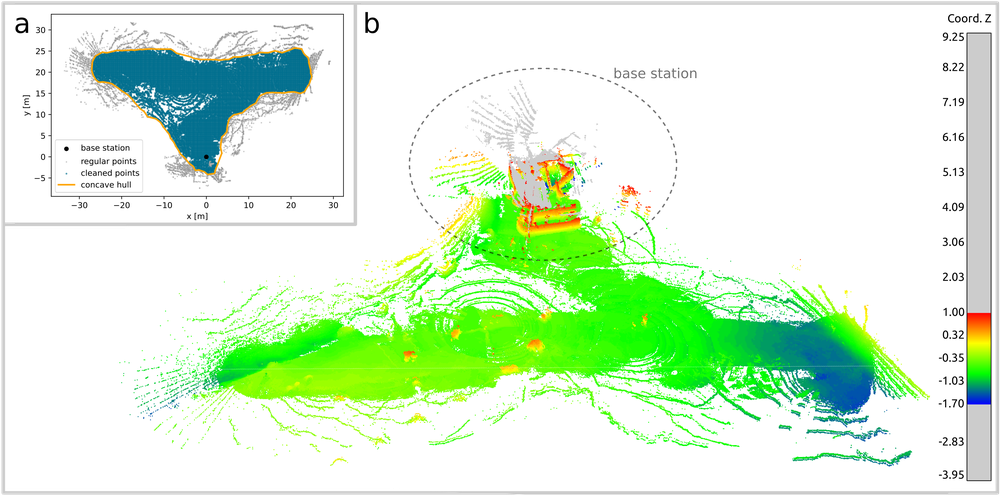}
    \caption{
    Multi-robot mapping results.
    a) Aggregated point cloud (\textcolor[HTML]{959595}{grey}) with concave hull (\textcolor[HTML]{FFA500}{orange}) outlining the mapped area used for area computation (\textcolor[HTML]{007091}{blue}).
    b) Assembled point cloud colored by height.}
    \label{fig:final_vdb}
\end{figure*}

\subsection{Exploration and Mapping Performance}
This section evaluates the exploration behavior and mapping performance of the robotic team, focusing on spatial coverage, mapping efficiency, and map accuracy achieved during the mission.
The traveled paths for each robot during the final 40-minute mission are shown in \cref{fig:mission_paths:paths}. 
The two remaining scout robots, Spot and \anontext{Dilly}{ANYmal~2}, initially explored different directions from the lander before converging toward the primary mission area.
The final assembled map is shown in \cref{fig:final_vdb}. 
After downsampling to a \SI{0.5}{\meter} resolution and extracting a concave hull (Figure~\ref{fig:final_vdb}a), the mapped area amounts to \SI{758}{\meter\squared}.
The scouts walked \SI{216}{\meter} and \SI{130}{\meter}, resulting in an overall mapping efficiency of \SI{2.19}{\square\meter\per\meter}.
With a mission duration of \SI{40.3}{\minute}, the \textit{Mapping Rate} was \SI{0.313}{\square\meter\per\second}.
Given the size of the mission area, the utilization of multiple \glspl{ugv} improved the~\textit{Total Explored Area} compared to using a single \gls{ugv}.
\par
To analyze the \textit{Map Error}, we compared the assembled point cloud to a ground truth map of the full area, generated by a single robot under ideal conditions beforehand.
We first cropped parts of the assembled cloud outside the ground truth coverage (≈10-15\%) and then computed the cloud-to-cloud Chamfer distances to the reference map, resulting in a mean of \SI{0.145}{\meter} and a standard deviation of \SI{0.252}{\meter}.
A substantial portion of the error originates from the region around mission control, highlighted in Figure \ref{fig:final_vdb}, where human safety operators interfered with mapping.
Overall, it can be observed that the robots only picked up high-fidelity information within a radius of \approx \SI{2}{\meter}, with only rough outlines being registered at greater distances.

\subsection{Scientific Measurement Performance}
This subsection assesses scientific measurement performance using resource identification ratios, execution success, and metrics for data quantity and quality.
A core task for scouts was identifying promising measurement positions for scientist robots.
As mentioned before \anontext{Dodo}{ANYmal~1}, the robot equipped with automatic rock detection could not take part in the mission due to water damage.
Therefore, the ratio of autonomously identified rocks could not be determined.
Identification of potential resources was thus performed by the human team, based on map and image data collected by the robots.
We classify identified resources into two categories. 
A normal measurement requires the target to appear in the map (rock targets only) and be clearly visible in one of the robots' camera feeds.
A scientific measurement additionally includes a successful measurement with either the RFA or the microscope.
For normal measurements, we identified 14 of 26 rocks and 4 of 6 sand patches, corresponding to identification ratios of 53.8\% and 66.7\%, respectively, and a total \textit{Ratio of Identified Resources} of 56.2\%.
Relevant scientific measurements were conducted on four rocks and three ground patches, yielding a measurement ratio of 21.8\%.
Both ratios were primarily constrained by the overall mission's time constraints.
\par
As for the scientists, \anontext{Donkey}{ANYmal~3} performed targeted inspections based on operator-selected POIs and successfully executed multiple scientific measurements, while
Husky performed several ground measurements near the lander area.
Due to high penalties assigned to prolonged motion — introduced based on observed slippage in the days preceding the mission — Husky’s autonomous behavior favored \glspl{poi} close to its current position.
Some measurements required multiple attempts to complete a \gls{poi}, with the \textit{Retry Ratio} quantifying the proportion of failed attempts relative to the total number of \gls{poi} attempts.
For \anontext{Donkey}{ANYmal~3}, two retries were required to complete 12 ground and rock \glspl{poi}, resulting in a \textit{Retry Ratio} of 16.6\% (2 failed attempts out of 14 total).
Husky completed seven ground-measurement \glspl{poi} with two retry attempts and one failure, yielding a \textit{Retry Ratio} of 22.2\% (2 retries over 9 total attempts).
Overall, this corresponds to an average \textit{Retry Ratio} of 19.4\%.
Concerning the \textit{Task Success Ratio}, Husky successfully completed 6 of 7 ground measurements (85.7\%), while \anontext{Donkey}{ANYmal~3} achieved a 100\% for its five ground measurements.
For the more challenging rock measurements, \anontext{Donkey}{ANYmal~3} succeeded in 3 out of 5 attempts (60\%), resulting in an overall measurement \textit{Success Ratio} of 82.3\%.
The high success ratios indicate that the scientists' autonomous measurement behavior is robust enough to withstand small disturbances.
The failures primarily occur when the robot is not positioned close enough to the measurement location, resulting in unreliable contact with the measurement surface.
\begin{figure*}[t]
    \centering
    \includegraphics[width=\textwidth]{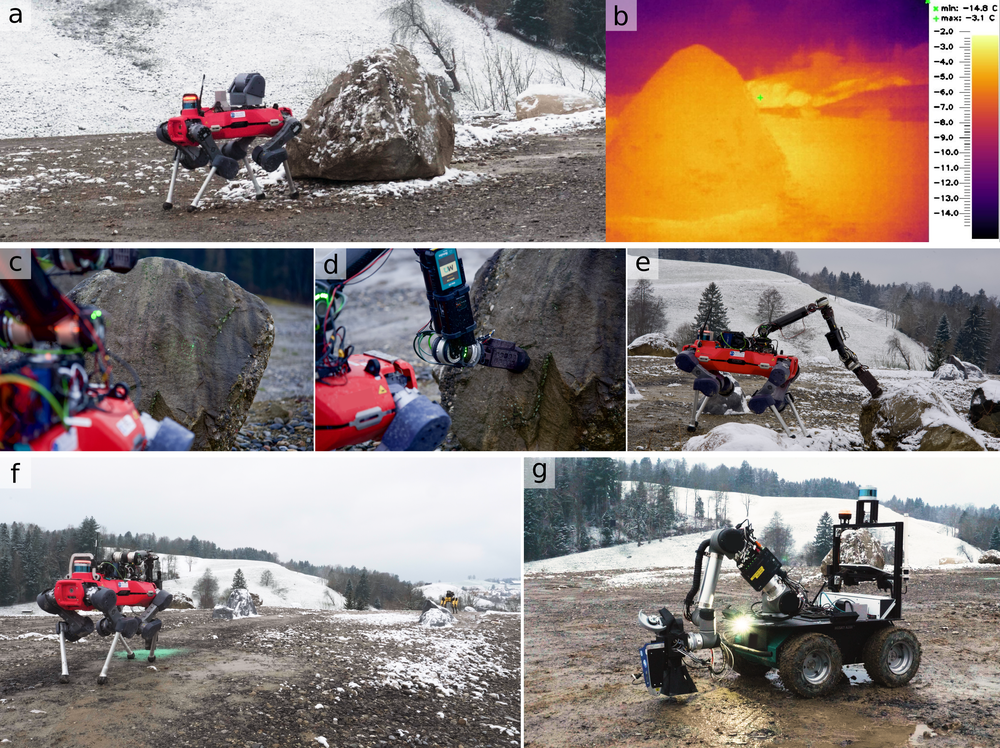}
    \caption{
        Representative scientific inspection and measurement tasks performed during field experiments.
        (a) \anontext{Dilly}{ANYmal~2} inspecting a rock target using a thermal camera.
        (b) Corresponding thermal image of the same target.
        (c-e) \anontext{Donkey}{ANYmal~3} performing rock measurement with the RAMAN spectrometer (c) and the microscope (d-e).
        (f) \anontext{Donkey}{ANYmal~3} performing a ground measurement with the microscope on a green sand path.
        (g) Husky performing an XRF ground measurement.
    }
    \label{fig:mission_measurements}
\end{figure*}

\begin{figure*}[t]
    \centering
    \includegraphics[width=\textwidth]{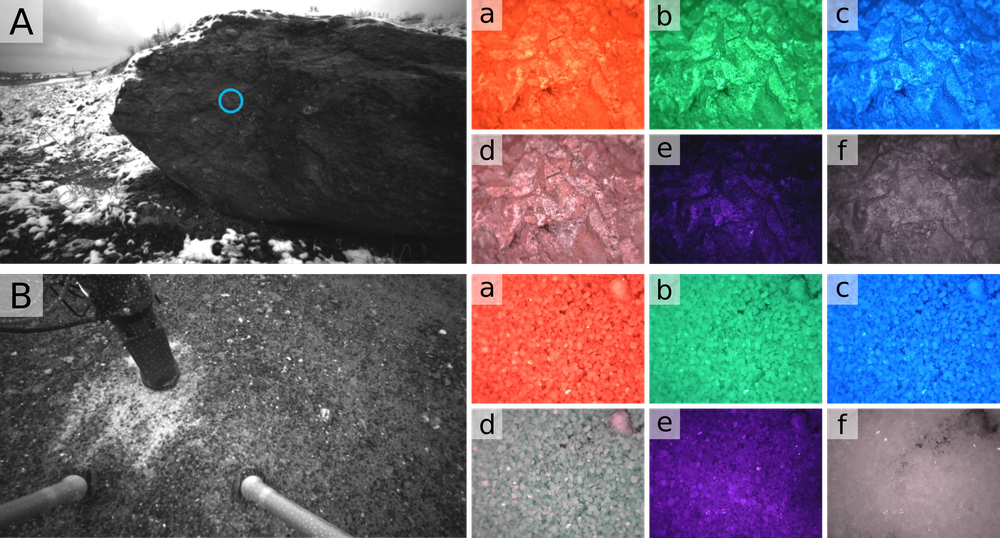}
    \caption{
    Example microscope measurements. (A) Rock measurement: wide-angle camera image of a rock with the measurement location highlighted in \textcolor[HTML]{4575b4}{blue}. (B) Ground measurement: infrared image from the onboard stereo camera showing blue sand. For each measurement, microscope images (a–f) illustrate the same sample under different illumination: (a) red, (b) green, (c) blue, (d) white, (e) UV, and (f) infrared light.
    }
    \label{fig:donkey_results}
\end{figure*}
\par
\par
The quality of the scientific measurements was assessed by scientists participating in the mission.
Husky’s RFA scan was largely impaired by the wet soil conditions, as the instrument cannot reliably detect light elements such as H, C, N, and O.
As a result, the spectra were dominated by undetectable light elements (≈85\%), offering little interpretive value.
By contrast, although Donkey’s Raman spectrometer delivered blurry spectra due to suboptimal tuning, the robot obtained a substantial set of high-quality microscope measurements.
These results, several of which are shown in \Cref{fig:donkey_results}, provide detailed and reliable visual information on the sampled materials.
\par
Overall, the team continued to operate effectively despite short-term hardware failures and challenging environmental conditions.
Roles were redistributed as needed, and the heterogeneous mix of platforms helped maintain mission capability even when individual robots could no longer perform their intended functions.
This adaptability reflects the underlying principles of our~\mosaic framework, where robustness to individual failures and flexible task reallocation are essential design goals.

%% file: sections/07_lessons_learned.tex
\section{LESSONS LEARNED}
\label{sec: LESSONS LEARNED}
To advance future multi-robot exploration systems, we provide a concise overview of the lessons learned from our field tests.
While not exhaustive, these insights should outline core principles relevant to most multi-robot systems.
\paragraph{Robot Interoperability}
While, on a technical level, multiple robot systems can be integrated using standardized interfaces and middleware, such as \gls{ros2}, in practice this is seldom straightforward.
The wide variety of robot capabilities and interfaces makes conforming to a fixed interface set challenging in practice.
Having mostly homogeneous software stacks for both mission-level and robot-level components, allowing only variations at the driver level, would have significantly reduced integration effort and increased system stability.
\par
On a technical level, integrating~\gls{ros1} and~\gls{ros2} components is unstable and not recommended.
While technically possible, it introduced delays, timing issues, and compatibility problems that greatly affected productivity.
Especially, maintaining a consistent~\texttt{TF-Tree} across both ROS versions was challenging due to its tight timing constraints and implementation differences, leading to incompatible behavior.
Especially with the mission control's overarching~\texttt{TF-Tree}, which collects and combines~\texttt{TF transforms} from all robots, invalid tree structures or timing inconsistencies between the received transforms occurred frequently.
This was due to~\gls{ros2}~\gls{qos} settings for~\texttt{TF} that correctly require \texttt{reliable} communication but, in practice, lead to significantly increased bandwidth usage, resulting in package loss and high latency.
More details on this are outlined in the next paragraph on networking for multi-robot systems.
\par
For future iterations of~\mosaic, we aim to fully transition to~\gls{ros2} or, where not possible, encapsulate the remaining \gls{ros1} components behind a non-ROS interface to improve interoperability.
Additionally, we opt not to transmit any transform information between robots, instead relaying the information utilizing poses (e.g., relative position of the robot in its map frame, position of the robot's map in the global earth frame) and low-frequency joint state information to assemble consistent but separate \texttt{TF-Trees} on mission control and each of the robots.
Test missions have shown that this reduces perceived latency and network load, allowing for transmissions even in low-quality network areas.
\paragraph{Networking}
Another major point of contention was the communication and networking between robots.
Specifically, we experienced communication issues over wireless connections using~\gls{ros2} DDS middleware (specifically~\texttt{rmw\_fastrtps\_cpp}).
As previously discussed, the use of \texttt{reliable} \gls{qos} settings resulted in significant latency, which also affected topics configured with other \gls{qos} profiles.
Although the exact cause—network utilization or DDS internals—could not be conclusively identified, these issues significantly affected the communication design and required adaptation.
Our partitioning of the network into three \enquote{regions} (robot $\leftrightarrow$ mission control station, mission control station $\leftrightarrow$ science operations station, and internal communication) and the use of~\gls{ros2} domain bridges for separation helped significantly reduce communications-related issues.
For communication links involving wireless transmission between a robot and mission control, we strongly recommend using \texttt{best\_effort} and \texttt{volatile} topics and relying on services that require synchronous behavior.
Reducing the amount of~\texttt{reliable} DDS transmissions on these connections also had a positive impact.
For wired (mission control station $\leftrightarrow$ science operations station) or loopback (internal communication) connections, this is not needed.
These optimizations result in an average bandwidth requirement of \SI[per-mode=symbol]{\approx 20}{\mega\bit\per\second}, which is sufficient for maintaining full communication with the robot team over a large area while relying solely on \SI{2.4}{\giga\hertz} WiFi from a single access point.
One important lesson from these field tests was that identifying the optimal target for network utilization in the final deployment area should be done quite early, as optimizing the networking architecture will substantially shape the system's design.
\paragraph{Co-Localization}
Due to the small scale of the test mission performed, we opted only for an initial localization of the robots against the start point at the mission control station.
During mission runtime, the robots would not re-localize against each other, allowing incremental misalignments due to drift.
Given the relatively small map sizes and short mission durations, this did not cause substantial misalignments.
For longer deployments or missions spanning large areas, a reliable co-localization strategy is needed.
By using online inter-robot map registration and smaller submaps, robots could account for continuous odometry drift and localization errors.
Addressing this limitation is a key future work.
\paragraph{Team Composition}
Upon reviewing the team composition for the field test, we identified several key takeaways for future missions.
First, having redundancies offers real benefits in terms of failure tolerance and parallelization.
Our approach has unexpectedly shown that compensating for a robot's full failure shortly before mission start is possible with minimal overhead, at the expense of the autonomy ratio.
\par
Secondly, we recommend maintaining a relatively high ratio of scouts to scientists (at least $3:2$).
Covering larger areas quickly allows scientists to split up and calculate more optimal paths for fulfilling their scientific goals.
However, when the ratio is too small, scientists will follow the scouts directly, reducing some of the advantages of having distinct roles.
Another advantage is that the exploration frontier size increases non-linearly with respect to the already explored area, thus requiring more area to be covered by each robot the longer the mission runs.
\par
Lastly, we highly recommend legged robots for exploration or science operations.
Their increased mobility allows them to traverse obstacles much more easily and safely than wheeled platforms.
While wheeled platforms are statically stable at all times, the dynamic nature of legged systems makes them much more adaptable to terrain, thereby removing operating-area restrictions.
They also benefit from higher speeds on rough terrain.
With the increased availability of manipulators on legged platforms, they can perform manipulation-dependent tasks in a manner almost identical to wheeled platforms, even in terms of stability.
Battery life remains a limitation for legged systems compared to wheeled systems, requiring a charging mechanism for missions exceeding an hour.
\paragraph{Operator Workload}
While~\mosaic overall was able to cope with the loss of a scout robot, an increased operator workload was still noticeable.
Even though the \textit{Quantitative Operator Workload} was \approx 78\%, we would still argue that, for prolonged missions, this value should decrease further to ensure that, during failures or measurements, the operator remains attentive.
One possible improvement in this regard is the use of an automatically managed operator task queue, allowing the operator to focus on the task itself instead of determining the currently most relevant open task.
\par
Another major contributor to the operator's workload is manual teleoperation.
As these tasks require the operator to perform high-precision movements while maintaining awareness of other, possibly more urgent tasks, we recommend having an additional dedicated teleoperator.
While this increases the overall number of people required, it could allow a significantly larger team to be managed without significantly increasing the per-person workload.
Similar to our current science operations station operator, this specialized operator can provide additional expertise while still allowing the main operator to take over during periods of high workload.
\par
In general, the importance of accounting for and improving the operator's role in the systems can be highlighted.
Effectively utilizing this resource can significantly enhance the team's effectiveness, particularly in unexpected situations.
We see this as a major topic for future work on human-supervised autonomous robot teams.

%% file: sections/08_conclusion.tex
\section{CONCLUSION \& FUTURE WORK}
\label{sec: CONCLUSION}
In this work, we present~\textbf{\mosaic, a modular and scalable autonomy framework} that enables unified mission-level coordination of heterogeneous robotic teams.
The key contribution is a unified mission abstraction, based on system-level objectives and robot-specific tasks, and layered autonomy, enabling robots to allocate and execute exploration and measurement tasks in parallel while a single operator retains the ability to intervene at mission-, robot-, and driver-level to account for unexpected situations.
Our approach provides an operator with a simple interface for performing actions, ranging from direct teleoperation to running complex action sequences via behavior trees, without creating excessive workloads.
We demonstrated our system in a 40-minute field deployment with five robots in a lunar-analog scenario, showing robust mission execution and continued operation despite the complete loss of one platform. 
Finally, we distilled practical lessons learned—particularly around \gls{ros2} interoperability, wireless communication, and \gls{qos}, team composition, and operator workflows—to inform future multi-robot planetary missions.
\par
In future work, we will address open points on our~\mosaic approach highlighted in~\Cref{sec: LESSONS LEARNED}.
We aim to push our approach as close as possible to real deployments of lunar exploitation teams and other similar mission scenarios.
Specifically, we plan to achieve autonomous operation on the mission level during longer communication blackouts, between the robots and the mission control stations, by incorporating mesh communication between robots and decentralized task/\gls{poi} management.
This decoupling from a central authority while maintaining eventual consistency promises higher uptime and fault tolerance.
Additionally, we plan to extend operator management to reduce task management workload and enable the incorporation of additional specialized operators.
The goal is to optimize this system to manage larger teams (\textasciitilde 10 robots) with 2-3 operators.
Another planned addition is an extension of the objective and task management to enable tighter cooperation between robots for specific tasks.
Especially, forming ad hoc sub-teams for complex objectives and enabling closely coupled task execution are intended here.
Finally, we aim to provide a standardized interface for integrating~\gls{ros2}-based robots into the team framework, simplifying the integration of new systems into our~\mosaic architecture.

%% file: sections/xx_acknowledgments.tex
\section*{ACKNOWLEDGMENT}
\Cref{fig:robot_team} has been cosmetically improved using generative AI–based image editing tools. 
Specifically, human bystanders were removed from the background in images (a) and (d) for visualization clarity.

\ifanonymous
\else
The authors thank Marco Trentini, Gabriela Ligeza, Florian Kehl, Manthan Patel, and Gaspard Smith-Vaniz for their support during the field experiments.
They also thank Carina Obrecht and Martina Overbeck for their support with photography and documentation.
Furthermore, the authors thank Florian Kehl, Gabriela Ligeza, and Valentin T. Bickel for scientific support in setting up the lunar geology scenarios.
\fi

%% file: sections/yy_biography.tex
\begin{IEEEbiography}[{\includegraphics[width=1in,height=1.25in,clip,keepaspectratio]{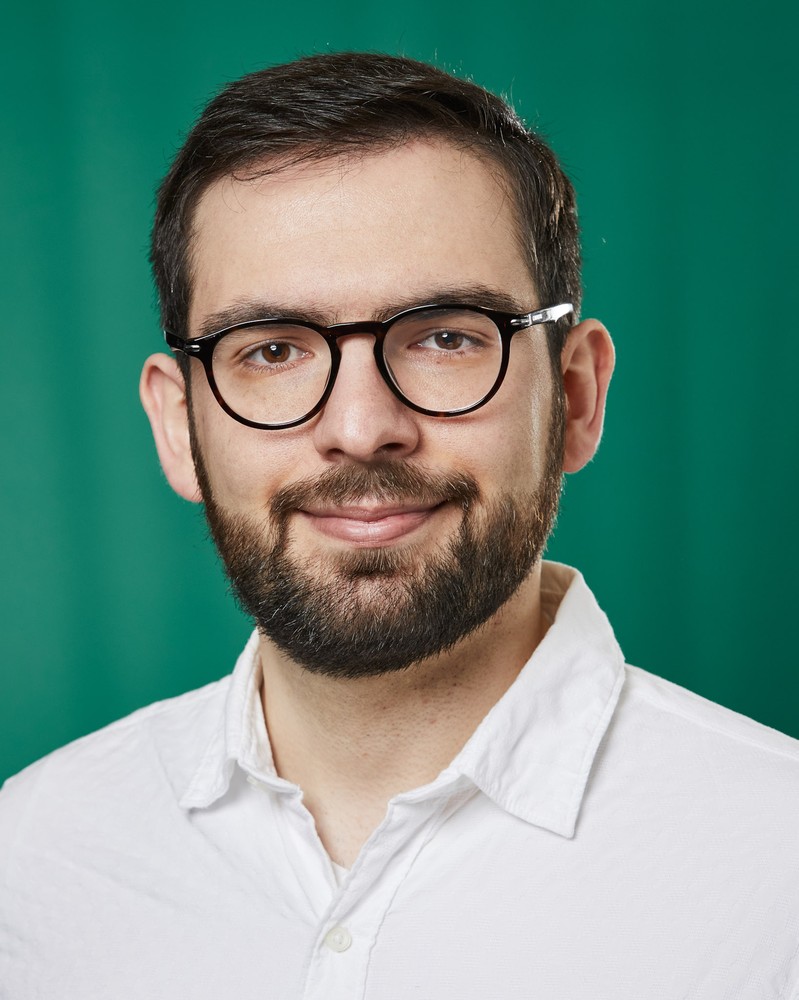}}]
    {\textsc{David Oberacker}}~received the M.Sc. degree in informatics from the Karlsruhe Institute of Technology (KIT), in Karlsruhe, Germany, in 2023, where he is currently pursuing the Ph.D. degree with the Machine Intelligence and Robotics Lab (MAIRO).

    He is with the FZI Research Center for Information Technology, Karlsruhe.
    His research interests focus on multi-robot coordination, shared semantic mapping, and multi-robot networking for field robotics.
\end{IEEEbiography}

\begin{IEEEbiography}[{\includegraphics[width=1in,height=1.25in,clip,keepaspectratio]{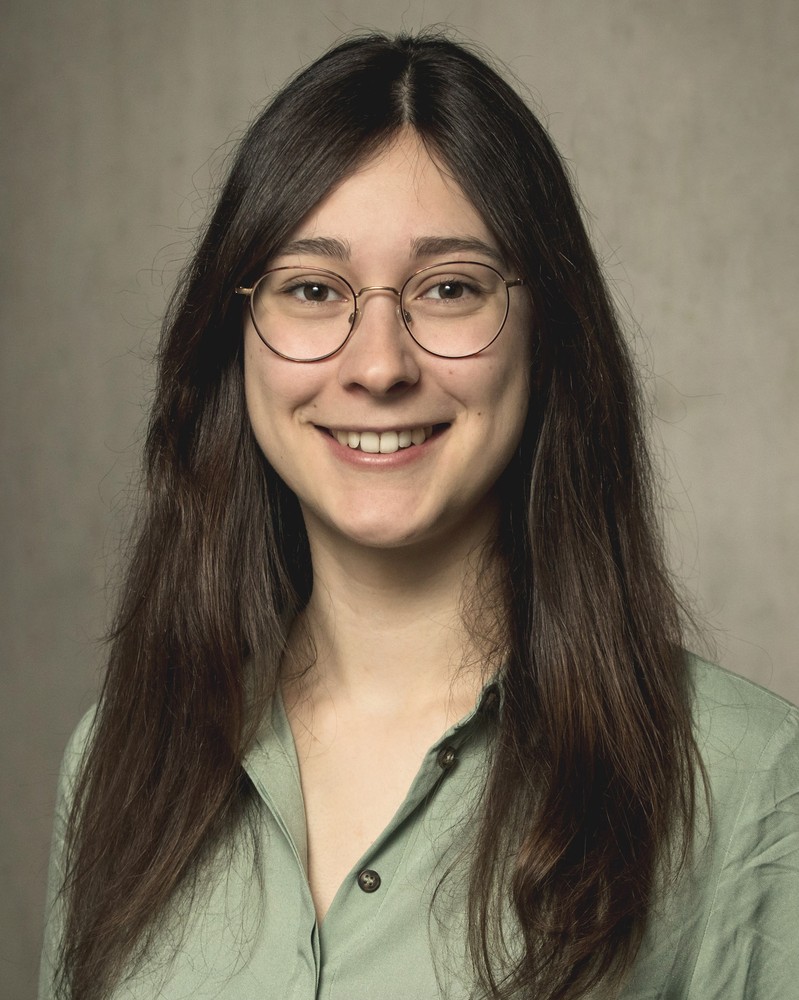}}]
    {\textsc{Julia Richter}}~received the Dipl.-Ing. degree in mechatronics from the Dresden University of Technology (TUD), Dresden, Germany, in 2023, and is currently pursuing the Ph.D. degree at the Robotic Systems Lab (RSL) at the Swiss Federal Institute of Technology (ETH Zürich), Zürich, Switzerland.

    Her research interests include autonomous navigation in unstructured environments, global path planning under uncertainty, and the integration of local perception with global map reasoning for long-range robotic mobility.
\end{IEEEbiography}

\begin{IEEEbiography}[{\includegraphics[width=1in,height=1.25in,clip,keepaspectratio]{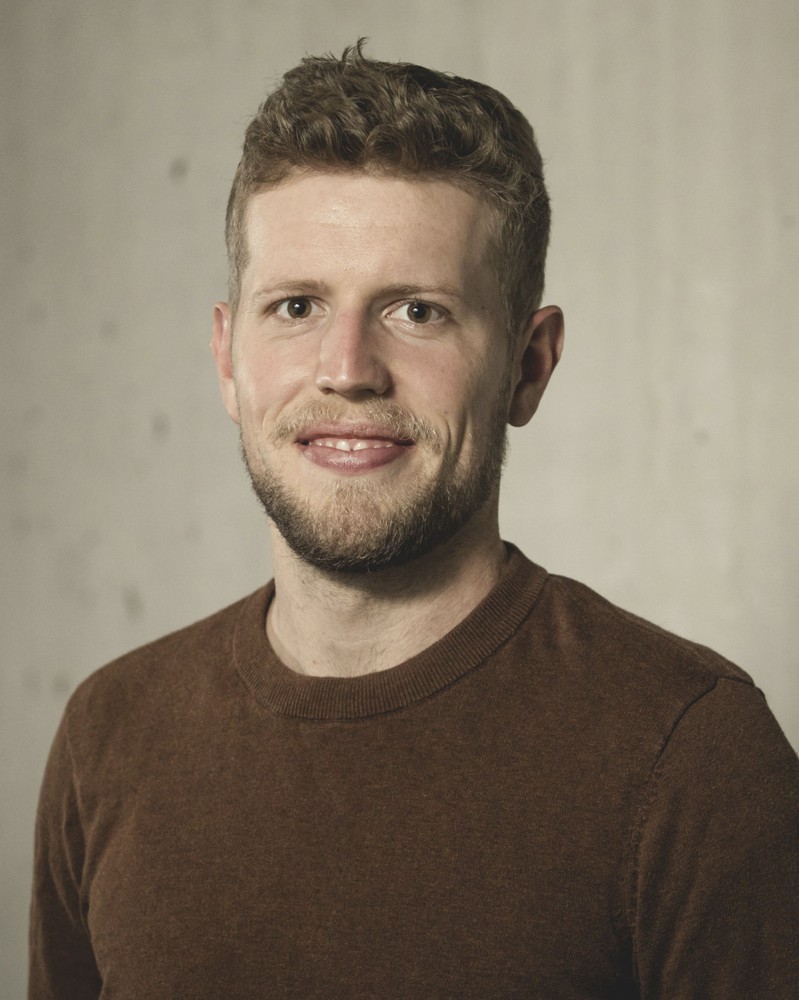}}]
    {\textsc{Philip Arm}}~received the M.Sc. degree in Robotics, Systems, and Control from ETH Zurich, Switzerland.
    He completed his Ph.D. in 2025 at the Robotic Systems Lab (RSL) at ETH Zurich, focusing on the development of legged robots for planetary exploration.
    He also completed an internship at the European Space Agency’s (ESA) Robotics and Automation Section, where he worked on locomotion strategies for legged robots in low-gravity environments. 
    
    His research focuses on locomotion and manipulation control of legged robots for planetary exploration and validation tests in extreme analog environments. 
\end{IEEEbiography}

\begin{IEEEbiography}[{\includegraphics[width=1in,height=1.25in,clip,keepaspectratio]{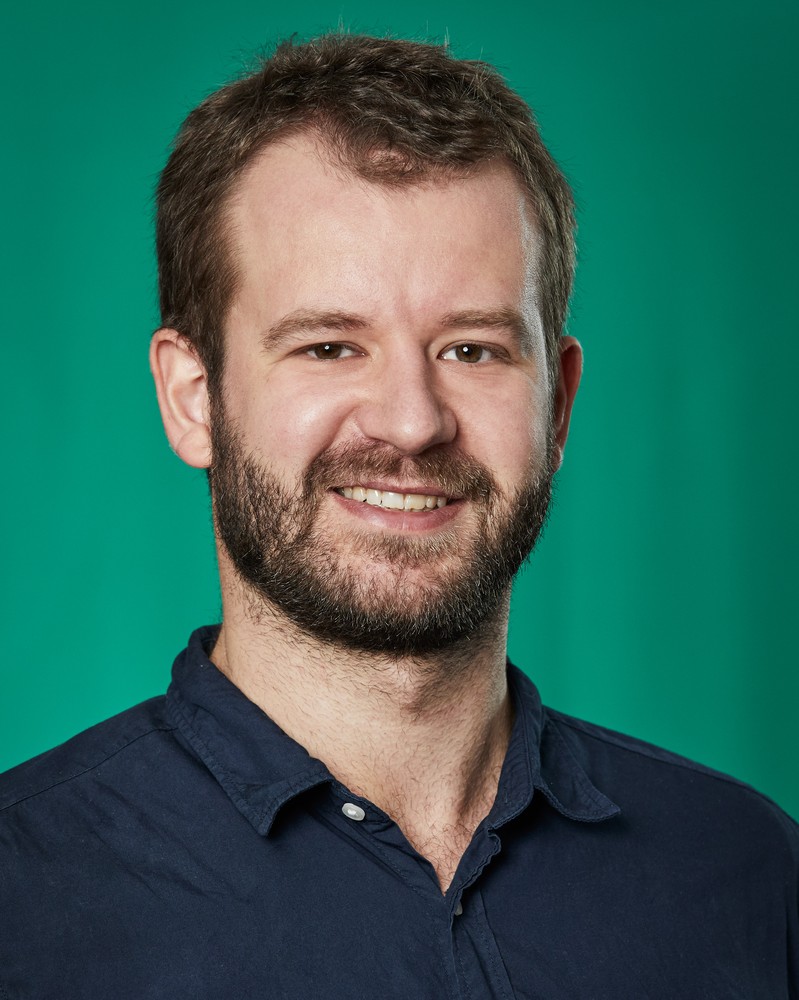}}]
    {\textsc{Lennart Puck}}~(Member, IEEE)~received the M.Sc. degree in computer science from the Karlsruhe Institute of Technology (KIT), Germany, in 2018, and the Ph.D. degree in 2024 from KIT, where his research focused on risk-aware exploration of unknown environments. From 2018 to 2024, he was with the FZI Research Center for Information Technology as a Researcher in the field of mobile robotics. 
    
    He is currently a Research Fellow in the Robotics Section at the European Space Agency (ESA), where he leads the Planetary Robotics Laboratory and works on autonomous robotic systems for space and planetary surface exploration.
\end{IEEEbiography}

\begin{IEEEbiography}[{\includegraphics[width=1in,height=1.25in,clip,keepaspectratio]{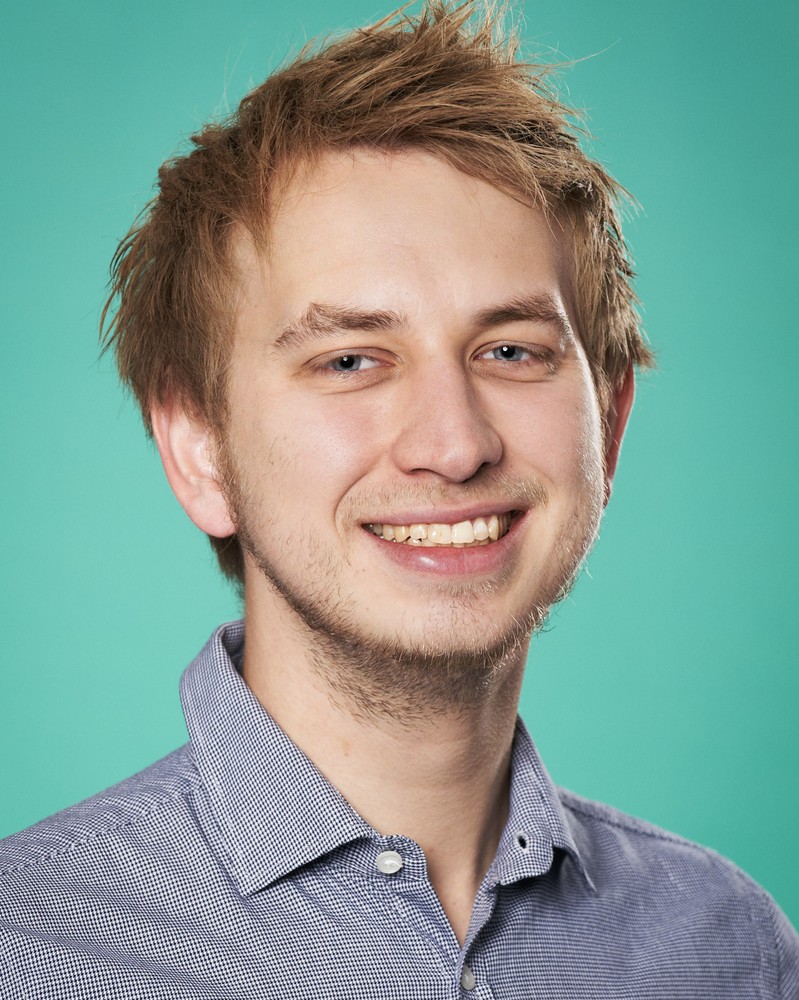}}]
    {\textsc{Marvin Grosse Besselmann}}~received the M.Sc. degree in computer science from the University of Lübeck, Germany. 
    In 2024, he completed his Ph.D. at the Karlsruhe Institute of Technology (KIT) in the domain of autonomous navigation on three-dimensional map representations.
    
    He is currently continuing his work as a Senior Researcher at the FZI Research Center for Information Technology, where his research focuses on state estimation, mapping, navigation, and long-term autonomy of heterogeneous robot teams.
\end{IEEEbiography}

\begin{IEEEbiography}[{\includegraphics[width=1in,height=1.25in,clip,keepaspectratio]{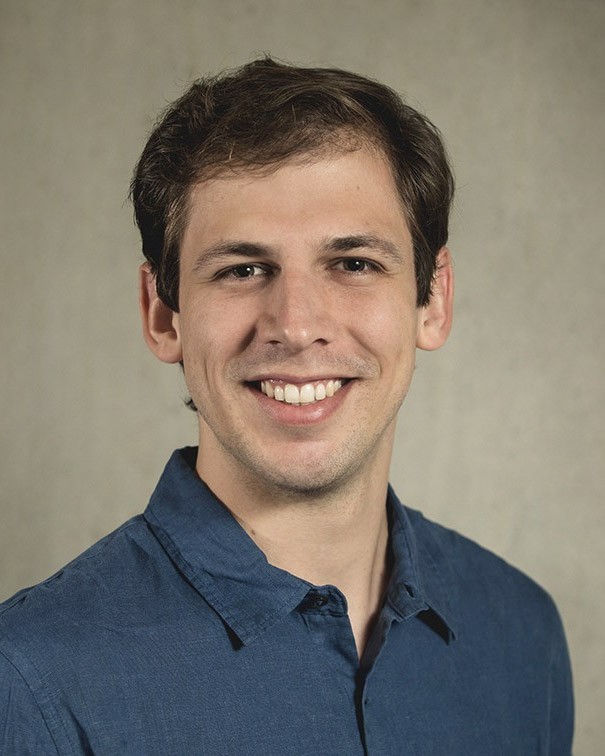}}]
    {\textsc{Will Talbot}}~is a PhD student in the Robotic Systems Lab at ETH Zurich.
    He received his M.Sc. in Robotics, Systems \& Control in 2024 from ETH Zurich. Previously, he completed a B.Eng in Mechatronic Engineering, majoring in Space Engineering, and a B.Sc, majoring in Computer Science and minoring in Physics, at the University of Sydney.
    His B.Eng thesis was conducted at the NASA Jet Propulsion Laboratory.
    
    His research interests include state estimation, mapping, and physical modeling.
\end{IEEEbiography}

\begin{IEEEbiography}[{\includegraphics[width=1in,height=1.25in,clip,keepaspectratio]{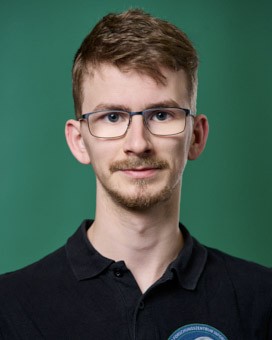}}]
    {\textsc{Maximilian Schik}}~received the M.Sc. degree in computer science in 2025 from the Karlsruhe Institute of Technology (KIT), in Karlsruhe, Germany.

    He is with the FZI Research Center for Information Technology.
    His research interests focus on robust locomotion on navigation.
\end{IEEEbiography}

\begin{IEEEbiography}[{\includegraphics[width=1in,height=1.25in,clip,keepaspectratio]{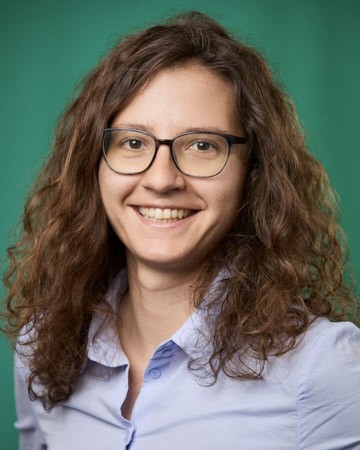}}]
    {\textsc{Sabine Bellmann}}~received the M.Sc. degree in mechanical engineering from the Karlsruhe Institute of Technology (KIT), in Karlsruhe, Germany.
    
    She is with the FZI Research Center for Information Technology in Karlsruhe, where her research focuses on the locomotion of legged robots on rough terrain.
\end{IEEEbiography}

\begin{IEEEbiography}[{\includegraphics[width=1in,height=1.25in,clip,keepaspectratio]{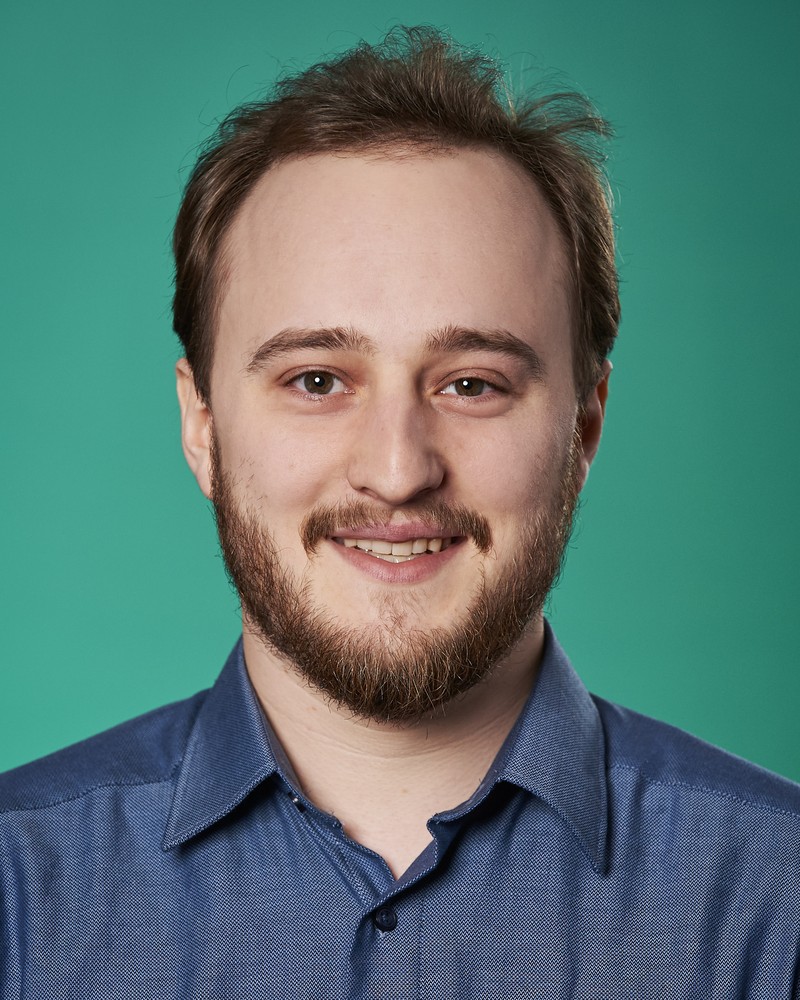}}]
    {\textsc{Tristan Schnell}}~received the M.Sc. degree in computer science in 2017 from the Karlsruhe Institute of Technology (KIT) in Karlsruhe, Germany.
    Since 2017, he has been working as a robotics researcher at the FZI Research Center for Information Technology.
    In 2024, he became head of the robotics lab.

    His research interests focus on robust autonomy for mobile robots, including robot self-awareness, multi-robot cooperation, and task and mission planning.
\end{IEEEbiography}

\begin{IEEEbiography}[{\includegraphics[width=1in,height=1.25in,clip,keepaspectratio]{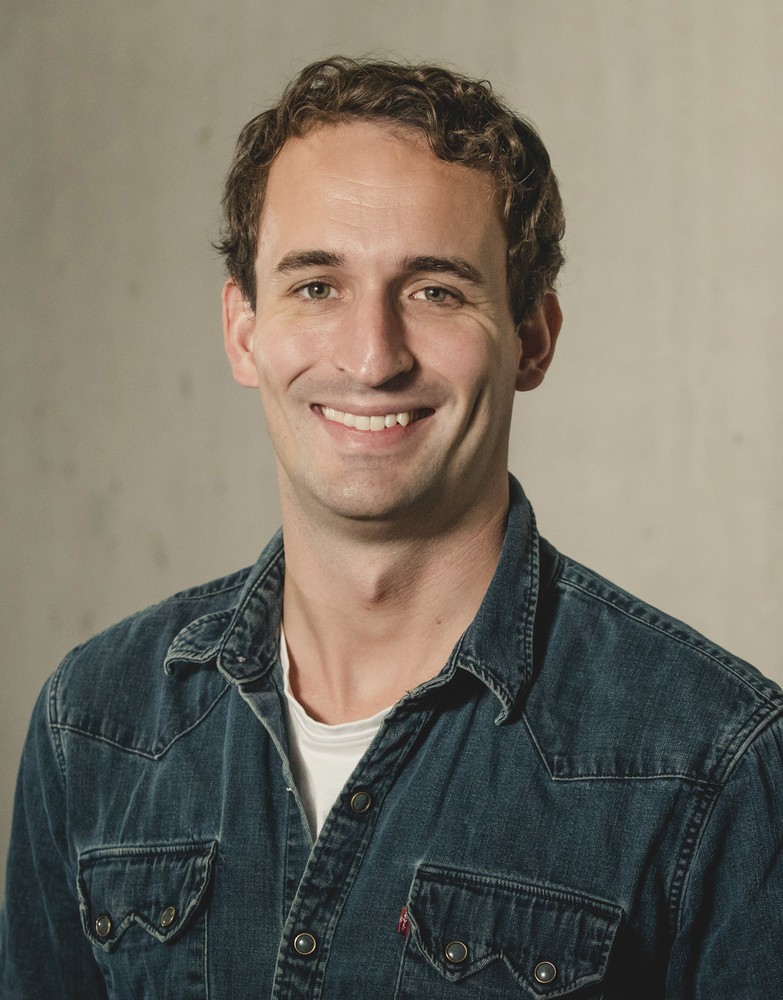}}]
    {\textsc{Hendrik Kolvenbach}}~received the B.Sc. and M.Sc. degrees in mechanical engineering from RWTH Aachen University, Aachen, Germany, in 2013 and 2014, respectively, and the Ph.D. degree in robotics from ETH Zürich, Zürich, Switzerland, in 2020, where he conducted his doctoral research with the Robotic Systems Lab (RSL).
    He is currently a Senior Researcher at RSL, ETH Zürich. 
    
    His research interests include robotic system design, legged locomotion, and the use of legged robots for planetary exploration and operation in challenging environments.
\end{IEEEbiography}

\begin{IEEEbiography}[{\includegraphics[width=1in,height=1.25in,clip,keepaspectratio]{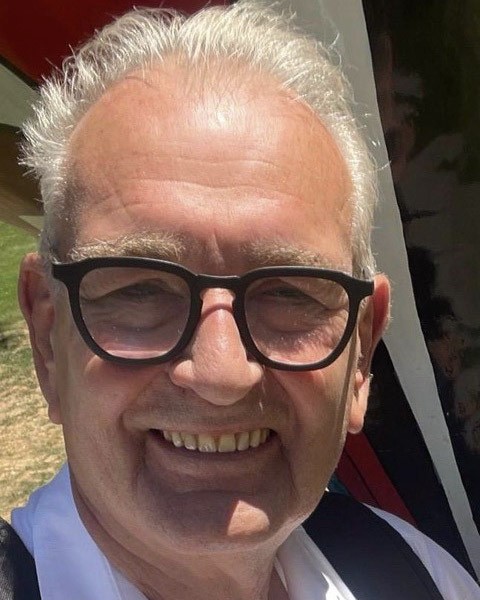}}]
    {\textsc{Rüdiger Dillmann}}~(Fellow, IEEE)~received his PhD from University of Karlsruhe in 1980.  
    Since 1987, he has been Professor of the Department of Computer Science and Director of the Humanoids and Intelligence Systems Lab at Karlsruhe Institute of Technology (KIT).
    In 2002, he became director of the innovation lab IDS (interactive diagnosis systems) at the Research Center for Information Science (FZI), Karlsruhe.
    In 2009, he founded the Institute of Anthropomatics and Robotics at KIT.
    
    His research interests are in the areas of human-robot interaction and neurorobotics, with special emphasis on intelligent, autonomous, and interactive robot behaviour generated with the help of machine learning methods and programming by demonstration (PbD).
    Other research interests include machine vision for mobile systems, man-machine cooperation, computer-supported intervention in surgery, and related simulation techniques.
    
    He is the author/co-author of more than 1000 scientific publications, conference papers, several books, and book contributions.
    He was the Coordinator of the German Collaborative Research Center “Humanoid Robots” and of several large-scale European IPs.
    He is the Editor-in-Chief of the book series COSMOS, Springer.
    Since 2018, he is Professor emeritus. 
    He is now the research director at FZI and is consulting start-up companies and SMEs of his former PhD students.
\end{IEEEbiography}

\begin{IEEEbiography}[{\includegraphics[width=1in,height=1.25in,clip,keepaspectratio]{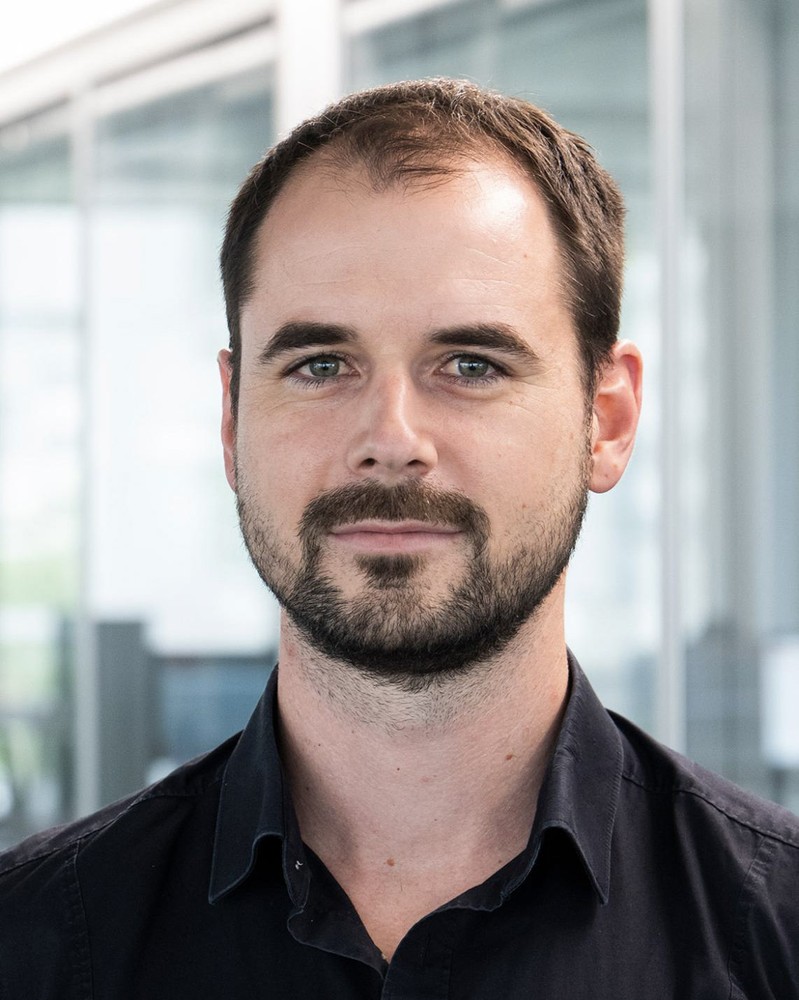}}]
    {\textsc{Marco Hutter}}~(Member, IEEE)~received the M.Sc. and Ph.D. degrees in design, actuation, and control of legged robots from the Swiss Federal Institute of Technology (ETH Zürich), Zürich, Switzerland, in 2009 and 2013, respectively.
    
    He is currently a Professor of robotic systems and the Director of the Center for Robotics, ETH Zürich. He is a Co-Founder of several ETH startups, such as ANYbotics in Zurich, Switzerland, and Gravis Robotics in Zurich. He is also the Director of Boston Dynamics AI Institute Zurich Office, Zurich. He is the Principal Investigator of the NCCRs robotics, automation, and digital fabrication. His research interests include the development of novel machines and actuation concepts, along with the underlying control, planning, and machine learning algorithms for locomotion and manipulation. Dr. Hutter was a recipient of the ERC Starting Grant and the winner of the DARPA SubT Challenge.
\end{IEEEbiography}

\begin{IEEEbiography}[{\includegraphics[width=1in,height=1.25in,clip,keepaspectratio]{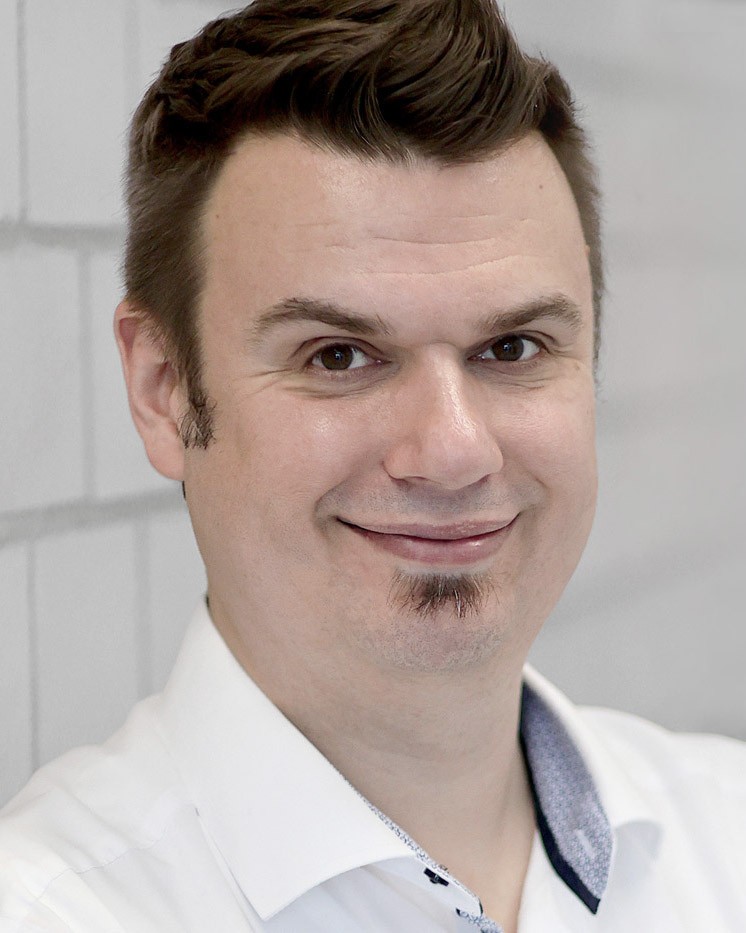}}]
    {\textsc{Arne Roennau}}~(Senior Member, IEEE) received his M.S. degree (Dipl.-Ing.) in electrical engineering and Ph.D. degree in computer science dedicated to the design, perception, and control of legged robots from the Karlsruhe Institute of Technology (KIT), Karlsruhe, Germany, in 2008 and 2019.
    
    He was a Postdoctoral Researcher and head of FZI Living Lab Service Robotics at the FZI Research Center for Information Technology. Currently, he is a Full Professor and Director of the Machine Intelligence and Robotics Lab (MaiRo) at KIT, Karlsruhe.
    He is also a Scientific Director of FZI Research Center for Information Technology, Karlsruhe, and a Principal Investigator of the Robotics Institute Germany (RIG). His research interests include design concepts for intelligent walking and service robots, situational and self-awareness, and learnable actions and decisions for advanced autonomy.
\end{IEEEbiography}